\pdfoutput=1 

\documentclass[graybox]{svmult}

\usepackage{mathptmx}       
\usepackage{helvet}         
\usepackage{courier}        
\usepackage{type1cm}        
\usepackage{makeidx}         
\usepackage{graphicx}        
\usepackage{multicol}        
\usepackage[bottom]{footmisc}

\usepackage{amsmath}
\usepackage{pdfsync} 
\usepackage{url}     
\usepackage[backref]{hyperref}
\usepackage[numbers]{natbib} 
\usepackage{rotating}
\usepackage{array}
\usepackage{xspace} 

\graphicspath{{figures/}} 

\makeindex             

\newcommand{\myvcenter}[1]{\ensuremath{\vcenter{\hbox{#1}}}}
\def\imagecenter#1{\myvcenter{\null\hbox{#1}}}

\newcommand{\threeD}{\mbox{3-D}\xspace} 
\newcommand{\twoD}{\mbox{2-D}\xspace} 

\begin{document}

\title*{Stereoscopic Cinema\thanks{This was published as chapter 2 in R\'emi Ronfard and Gabriel Taubin, editors, Image and Geometry Processing for 3-D Cinematography, volume 5 of Geometry and Computing, pages 11–51. Springer Berlin Heidelberg, 2010. ISBN 978-3-642-12392-4. doi: \href{http://dx.doi.org/10.1007/978-3-642-12392-4_2}{10.1007/978-3-642-12392-4\_2}. The original publication is available at \href{http://www.springerlink.com/}{www.springerlink.com}.}}
\author{Fr\'ed\'eric Devernay and Paul Beardsley}
\institute{Fr\'ed\'eric Devernay \at INRIA Grenoble - Rh\^one-Alpes, France \email{frederic.devernay@inria.fr} \and Paul Beardsley \at Disney Research, Z\"urich, \email{pab@disneyresearch.com}}
%
%
\maketitle

\abstract{Stereoscopic cinema has seen a surge of activity in recent years, and for the first time all of the major Hollywood studios released \threeD{} movies in 2009.  This is happening alongside the adoption of \threeD{} technology for sports broadcasting, and the arrival of 3-D TVs for the home.  Two previous attempts to introduce \threeD{} cinema in the 1950s and the 1980s failed because the contemporary technology was immature and resulted in viewer discomfort.  But current technologies -- such as accurately-adjustable \threeD{} camera rigs with onboard computers to automatically inform a camera operator of inappropriate stereoscopic shots, digital processing for post-shooting rectification of the \threeD{} imagery, digital projectors for accurate positioning of the two stereo projections on the cinema screen, and polarized silver screens to reduce cross-talk between the viewers left- and right-eyes -- mean that the viewer experience is at a much higher level of quality than in the past.  Even so, creation of stereoscopic cinema is an open, active research area, and there are many challenges from acquisition to post-production to automatic adaptation for different-sized display.  This chapter describes the current state-of-the-art in stereoscopic cinema, and directions of future work.  }

\section{Introduction}
\label{cha:introduction}

\subsection{Stereoscopic Cinema, \threeD{} Cinema, and Others}
\label{sec:ster-cinema-3}

Stereoscopic cinema\index{Stereoscopic cinema} is the art of making stereoscopic films or motion pictures. In stereoscopic films, depth perception is enhanced by having different images for the left and right human eye, so that objects present in the film are perceived by the spectator at different depths. The visual perception process that reconstructs the \threeD{} depth and shape of objects from the left and right images is called stereopsis.

Stereoscopic cinema is also frequently called \threeD{} cinema\footnote{Stereoscopic cinema is even sometimes redundantly called Stereoscopic \threeD{} (or S3D) cinema. }\index{3-D cinema|see{Stereoscopic cinema}}\index{Stereoscopic 3-D cinema|see{Stereoscopic cinema}}, but in this work we prefer using the term stereoscopic, as \threeD{} cinema may also refer to  other novel methods for producing moving pictures:
\begin{itemize}
\item  free-viewpoint video, where videos are captured from a large number of cameras and combined into 4-D data that can be used to render a new video as seen from an arbitrary viewpoint in space;
\item  \threeD{} geometry reconstructed from multiple viewpoints and rendered from an arbitrary viewpoint.
\end{itemize}
\citet{Smolic:2005} did a large review on the subjects of \threeD{} and free-viewpoint video, and other chapters in the present book also deal with building and using these moving pictures representations.

In this chapter, we will limit ourselves to movies that are made using exactly two cameras placed in a stereoscopic configuration. These movies are usually seen through a \threeD{} display device which can present two different images to the two human eyes (see \citet{Sexton:1999} for a review on \threeD{} displays). These display devices sometimes take as input more than two images, especially glasses-free \threeD{} displays where a number of viewpoints are displayed simultaneously but only two are seen by a human spectator. These viewpoints may be generated from stereoscopic film, using techniques that will be described later in this chapter, or from other \threeD{} representations of a scene.

Many scientific disciplines are related to stereoscopic cinema, and they will sometimes be referred to during this chapter. These include computer science (especially computer vision and computer graphics), signal processing, optics, optometry, ophthalmology, psychophysics, neurosciences, human factors, and display technologies.

\subsection{A Brief History of Stereoscopic Cinema}
\label{sec:short-history-3D}

The interest in stereoscopic cinema appeared at the same time as cinema itself, and the ``prehistory'' (from 1838 to 1922 when the first public projection was made~\cite{Gosser:1977,Zone:2007}) as well as ``history'' (from 1922 to 1952, when the first feature-length commercial stereoscopic movie, \emph{Bwana Devil}, was released to the public~\cite{Zone:2007}) follow closely the development of cinematography. The history of the period from 1952 to 2004 can be retraced from the thoughtful conversations found in \citet{Zone:2005}.

In the early 50's, the television was causing a large decrease in the movie theater attendance, and stereoscopic cinema was seen as a method to get back this audience. This explains the first wave of commercial stereoscopic movies in the 50's, when Cinemascope also appeared as another television-killer. Unfortunately the quality of stereoscopic movies, both in terms of visual quality and in terms of cinematographic content, was not on par with \twoD{} movies in general, especially Cinemascope. Viewing a stereoscopic film at that time meant the promise of a headache, because stereoscopic filming techniques were not fully mastered, and the spectacular effects were overly used in these movies, reducing the importance of screenplay to close to nothing. One notable counter-example to the typical weak screenplay of stereoscopic movies was \emph{Dial M for Murder} by Alfred Hitchcock, which was shot in \threeD{}, but incidentally had a much bigger success as a \twoD{} movie. Finally, Cinemascope won the war against TV, and the production of stereoscopic cinema declined.

The second stereoscopic cinema ``wave'', in the 80's, was formed both by large format (IMAX \threeD{}) stereoscopic movies, and by standard stereoscopic movies which tried again the recipe from the 50's (spectacular content but weak screenplay), with no more success than the first wave. There were a few high-quality movies, such as \emph{Wings of Courage} by Jean-Jacques Annaud in 1995, which was the first IMAX \threeD{} fiction movie, made with the highest standards, but at a considerable cost. The IMAX-3D film camera rig is extremely heavy and difficult to operate~\citep{Zone:2005}. Stereoscopic cinema wouldn't take off until digital processing finally brought the tools that were necessary to make stereoscopic cinema both easier to shoot and to watch.

As a matter of fact, the rebirth of stereoscopic cinema came from animation, which produced movies like \emph{Chicken Little}, \emph{Open Season}  and \emph{Meet the Robinsons}, which were digitally produced, and thus could be finely tuned to lower the strain on the spectator's visual system, and experiment new rules for making stereoscopic movies. Live-action movies produced using digital stereoscopic camera systems came afterwards, with movies such as \emph{U2 3D}, and of course \emph{Avatar}, which held promises of a stereoscopic cinema revival.

Stereoscopic cinema brought many new problems that were not addressed by traditional movie making. Many of these problems deal with geometric considerations: how to place the two cameras with respect to each other, where to place the actors, what camera parameters (focal length, depth of field, ...) should be used... As a matter of fact, many of these problems were somehow solved by experience, but opinions often diverged on the right solution to film in stereoscopy. The first theoretical essay on stereoscopic cinema was written by \citet{Spottiswoode:1952}. The Spottiswoodes made a great effort to formalize the influence of the camera geometry on the \threeD{} perception by the audience, and also proposed a solution on the difficult problem of transitions between shots.  Their scientific approach of stereoscopic cinema imposed very strict rules on moviemaking, and most of the stereoscopic moviemakers didn't think it was the right way to go~\cite{Zone:2005}.

Many years later, Lenny Lipton, who founded StereoGraphics and invented the CrystalEyes electronic shutter glasses, tried again to describe the scientific foundations of stereoscopic cinema~ \cite{Lipton:1982}. His approach was more viewer-centric, and he focused more on how the human visual system perceives \threeD{}, and how it reacts to stereoscopic films projected on a movie screen. Although the resulting book contains many mathematical errors, and even forgets most of the previous finding by the Spottiswoodes~\cite{Smith:1983}, it remains one of the very few efforts in this domain.

The last notable effort at trying to formalize stereoscopic movie making was this of \citet{Mendiburu:2009}, who omitted maths, but explained with clear and simple drawings the complicated geometric effects that are involved in stereoscopic cinema. This book was instantaneously adopted as a reference by the moviemaking community, and is probably the best technical introduction to the domain.

\subsection{Computer Vision, Computer Graphics, and Stereoscopic Cinema}
\label{sec:comp-visi-comp}

The discussions in this chapter straddle Computer Vision, Computer Graphics, and Stereoscopic Cinema: Computer Vision techniques will be used to compute and locate the defects in the images taken by the stereoscopic camera, and Computer Graphics techniques will be used to correct these defects.

\subsubsection{A Few Definitions}
\label{sec:few-definitions}

Each discipline has its own language and words. Before proceeding, let us define a few geometric elements that are useful in stereoscopic cinema~\cite{Hummel:2008}:

\textbf{Interocular (also called Interaxial):} (the term baseline is also widely used in computer vision, but not in stereoscopic cinema) The distance between the two eyes/cameras, or rather their optical centers. It is also used sometimes used to designate the segment joining the two optical centers. The average human interocular is 65mm, with large variations around this value.

\textbf{Hyperstereo (or miniaturization):}\index{Hyperstereo}\index{Miniaturization|see{Hyperstereo}} The process of filming with an interocular larger than 65mm (it can be up to a few dozen meters), with the consequence that the scene appears smaller when the stereoscopic movie is viewed by a human subject.

\textbf{Hypostereo (or gigantism):}\index{Hypostereo}\index{Gigantism|see{Hypostereo}} The process of filming with an interocular smaller than 65mm, resulting in a ``bigger than life'' appearance when viewing the stereoscopic movie. It can be as small as 0mm.

\textbf{Roundness factor:}\index{Roundness factor} Suppose a sphere is filmed by a stereoscopic camera. When displaying it, the roundness factor is the ratio between its apparent depth and its apparent width. If it is lower than 1, the sphere appears as a flattened disc parallel to the image plane. If it is bigger than 1, the sphere appears as a spheroid elongated in the viewer's direction. We will see that the roundness factor depends on the object's position in space.

\textbf{Disparity:} The difference in position between the projections of a \threeD{} point in the left and right images, or the left and right retinas. In a standard stereo setup, the disparity is mostly horizontal (the corresponding points are aligned vertically), but vertical disparity may happen (and has to be corrected).

\textbf{Screen plane:} The position in space where the display projection surface is located (supposing the projection surface is planar).

\textbf{Vergence, convergence, divergence :} the angle formed by the the optical axis of the two eyes in binocular vision. The optical axis is the half \threeD{} line corresponding to the line-of-sight of the center of the fovea. It can be positive (convergence) or negative (divergence).

\textbf{Plane of convergence:} The vertical plane parallel to the screen plane containing the point that the two eyes are looking at. If it is in front of the screen plane, then the object being looked at appears in front of the screen. When using cameras, the plane of convergence is the zero-disparity plane (it is really a plane if images are rectified, as will be seen later).


\textbf{Proscenium arch (also called stereoscopic window and floating window):} (Fig.~\ref{fig:proscenium})\index{Proscenium arch}\index{Stereoscopic window|see{Proscenium arch}} The perceived depth of the screen borders. If the left and right borders of the left and right images do not coincide on the screen, the proscenium arch is not in the screen plane. It is also called the stereoscopic window, since the 3D-scene looks as if it were seen through that \threeD{} window. As will be explained later, objects closer than the proscenium arch should not touch the left or right side of the arch.

\subsubsection{Stereo-Specific Processes}
\label{sec:ster-spec-proc}

The stereoscopic movie production pipeline shares a lot with standard \twoD{} movie production~\cite{Mendiburu:2009}. Computer vision and computer graphics tools used in the process can sometimes be used with no or little modification. For example, matchmoving\index{Match moving} (which is more often called \emph{Structure from Motion} or \emph{SfM} in Computer Vision) can take into account the fact that two cameras are taking the same scene, and may use the additional constraint that the two camera positions are fixed with respect to each other.

Many processes, though, are specific to stereoscopic cinema production and post-production, and cannot be found in \twoD{} movie production:

\begin{itemize}

\item \emph{Correcting geometric causes of visual fatigue} such as images misalignments, will be covered by the next two sections (\ref{cha:sour-visu-fatig} and \ref{cha:elim-vert-disp}).

\item \emph{Color-balancing left and right images}\index{Stereoscopic cinema!color balancing} is especially necessary when a half-silvered mirror is used to separate images for the left and right cameras and wide angle lenses are used: the transmission and reflection spectrum response of these mirrors depend on the incidence angle and have to be calibrated. This can be done using color calibration devices and will not be covered in this chapter.

\item \emph{Adapting the movie to the screen size (and distance)} is not as simple as scaling the left and right views: the stereoscopic display is not easily scalable like a \twoD{} display, because the human interocular is fixed  and therefore does not ``scale'' with the screen size or distance. A consequence is that the same stereoscopic movie displayed on screens of different sizes will probably give quite different \threeD{} effects (at least quantitatively). The adaptation can be done at the shooting stage (Sec.~\ref{cha:keep-prop-pick}) or in post-production (Sec.~\ref{cha:chang-shoot-geom}). These processes can either be used to give the stereoscopic scene the most natural look possible (which usually means a roundness factor close to 1), or to ``play'' with \threeD{} effects, for example by changing the interocular or the position where the infinity plane appears.

\item \emph{Local \threeD{} changes (or \threeD{} touchup)} consist in editing the \threeD{} content of the stereoscopic scene. This usually means providing new interactive editing tools that work both on the images and on the disparity map. These tools usually share a lot with colorizing tools, since they involve cutting-up objects, tracking them, and changing their depth (instead of color). This is beyond the scope of this chapter.

\item \emph{Playing with the depth of field} is sometimes necessary, especially when adapting the \threeD{} scene to a given screen distance: the depth of field should be consistent with the distance to the screen, in order to minimize vergence-accomodation conflicts (Sec.~\ref{sec:changing-depth-focus})

\item \emph{Changing the proscenium arch} is sometimes necessary, because objects in front of the screen may cross the screen borders and become inconsistent with the stereoscopic window (Sec~\ref{sec:break-prosc-rule}).

\item \emph{\threeD{} compositing (with real or CG scenes)}\index{Stereoscopic cinema!compositing}\index{Compositing|see{Stereoscopic cinema, compositing}} should be easier in stereoscopic cinema, since it has \threeD{} content already. However, there are some additional difficulties: \twoD{} movies mainly have to deal with positioning the composited objects within the scene and dealing with occlusion masks. In \threeD{}, the composited scene must also be consistent between the two eyes, and its \threeD{} shape must be consistent with the original stereoscopic scene (Sec.~\ref{sec:comp-ster-scen}). Relighting\index{Stereoscopic cinema!relighting}\index{Relighting!see{Stereoscopic cinema, relighting}} the scene also brings out similar problems.
\end{itemize}

\section{Three-Dimensional Perception and Visual Fatigue}
\label{sec:why-geometry-so}

In traditional \twoD{} cinema, since the result is always a 2D moving picture, almost anything can be filmed and displayed, without any effect on the spectator's health (except maybe for light flashes and stroboscopic effects), and the artist has a total freedom on what can be shown to the spectator. The result will appear as a moving picture drawn on a plane placed at some distance from the spectator, and will always be physically plausible.

In stereoscopic cinema, the two images need to be mutually consistent, so that a 3-D scene (real or virtual) can be reconstructed by the human brain. This implies strong geometric and photometric constraints between the images displayed to the left and the right eye, so that the \threeD{} shape of the scene is perceived correctly, and there is not too much strain on the human visual system which would result in visual fatigue.

Our goal in this chapter is to deal with the issues related to geometry and visual fatigue in stereoscopic cinema, but without any artistic considerations. We will just try to define bounds within which the artistic creativity can freely wander. These bounds were almost non-existent in \twoD{} movies, but as we will see they are crucial in stereoscopic movies.

Besides, since \threeD{} perception is naturally more important in stereoscopic movies, we have to understand what are the different visual features that produce depth perception. Those features will be called \emph{depth cues}\index{Depth cues}, and surprisingly most are monoscopic and can be experienced by viewing a \twoD{} image. Stereoscopy is only one depth cue amongst many others, although it may be the most complicated to deal with. The stereographer Phil Streather, quoting Lenny Lipton, said: ``Good 3D is not just about setting a good background. You need to pay good attention to the seven monocular cues -- aerial perspective, inter position, light and shade, relative size, texture gradients, perspective and motion parallax. Artists have used the first five of those cues for centuries. The final stage is depth balancing.''

\subsection{Monoscopic Depth Cues}
\label{sec:depth-cues}

\begin{figure}
  \centering
\begin{tabular}{|c|c|c|}
  \hline
\includegraphics[width=0.3\columnwidth]{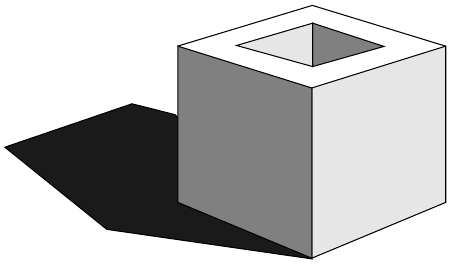} & \includegraphics[width=0.3\columnwidth]{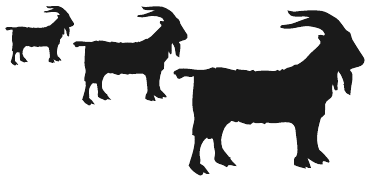} & \includegraphics[width=0.3\columnwidth]{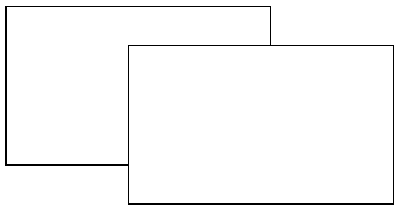} \\
Light and shade & Relative size & Interposition \\
\hline
\includegraphics[width=0.3\columnwidth]{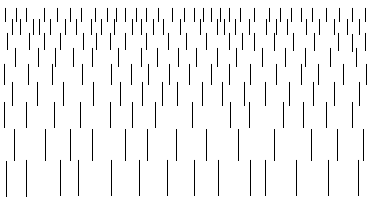} & \includegraphics[width=0.3\columnwidth]{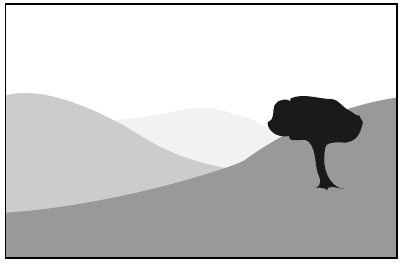} & \includegraphics[width=0.3\columnwidth]{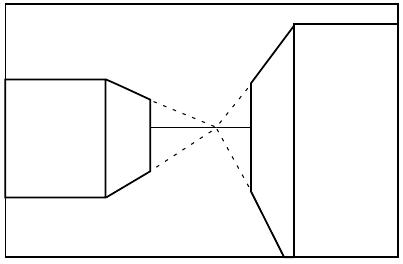} \\
Textural gradient & Aerial perspective & Perspective \\
\hline
\end{tabular}
  \caption{Six monoscopic depth cues (from \cite{Stereographics:1997}). The seventh is motion parallax, which is hard to illustrate, and depth of field can also be considered as a depth cue (see Fig.~\ref{fig:focusmatters}).}
  \label{fig:monoscopiccues}
\end{figure}

The perception of \threeD{} shape is caused by the co-occurrence of a number of consistent visual artifacts called depth cues. These depth cues can be split into monoscopic cues, and stereoscopy (or stereopsis). For a review of \threeD{} shape perception from the cognitive science perspective, see \citet{Todd:2004}. The basic seven monoscopic depth cues, as described in \citet[chap. 2]{Lipton:1982}  (see also \cite{Stereographics:1997}), and illustrated Fig.~\ref{fig:monoscopiccues}, are:
\begin{itemize}
\item \emph{Light and shade}\index{Light and shade}
\item \emph{Relative size}\index{Relative size}, or retinal image size size (smaller objects are farther)
\item \emph{Interposition}\index{Interposition}, or overlapping (overlapped objects lie behind)
\item \emph{Textural gradient}\index{Textural gradient} (increase in density of a projected texture as a function of distance and slant)
\item \emph{Aerial perspective}\index{Aerial perspective} (usually caused by haze)
\item \emph{Motion parallax}\index{Motion parallax} (\twoD{} motion of closer objects is faster)
\item \emph{Perspective}\index{Perspective}, or linear perspective
\end{itemize}

\begin{figure}[t]
  \centering
  \includegraphics[width=0.73\columnwidth]{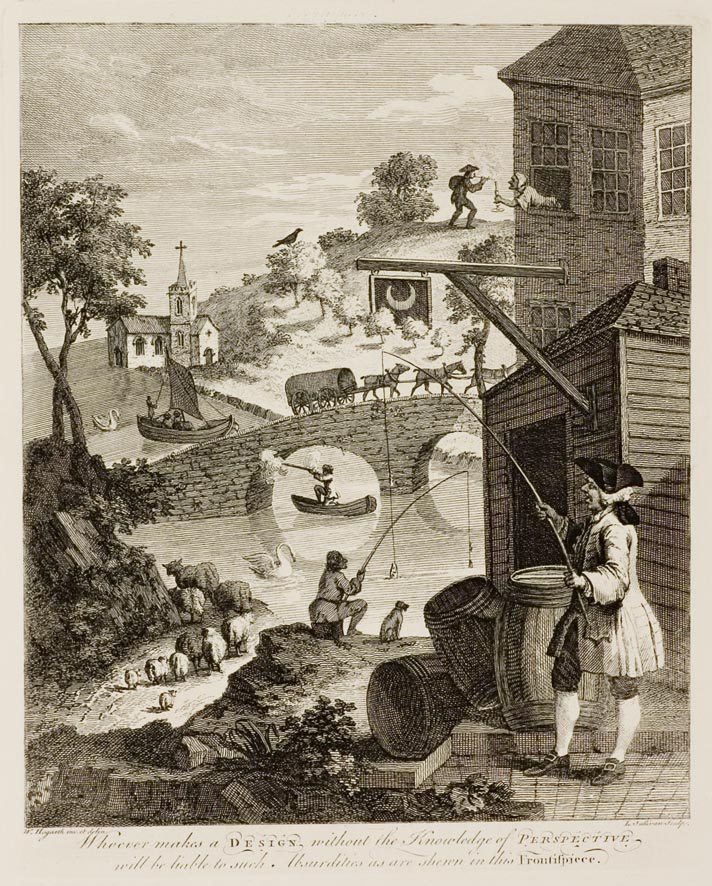}
  \caption{``Whoever makes a \textsc{design} without the Knowledge of \textsc{perspective} will be liable to such Absurdities as are shown in this Frontispiece'' (engraving by William Hogarth, 1754), a proof by contradiction of the importance of many monoscopic depth cues.}
  \label{fig:hogarth}
\end{figure}

As can bee seen in a famous drawing by Hogarth (Fig.~\ref{fig:hogarth}), their importance can be easily demonstrated by using contradictory depth cues.

\citet{Lipton:1982} also refers to what he calls a ``physiological cue'': \emph{Accommodation} (the monoscopic focus response of the eye, or how much the ciliary muscles contract to maintain a clear image of an object as its distance changes). However, it is not clear from psychophysics experiments whether this should be considered as depth cue, i.e. if it gives an indication of depth in the absence of any other depth cue.

\begin{figure}
  \centering
  \includegraphics[width=0.52\columnwidth]{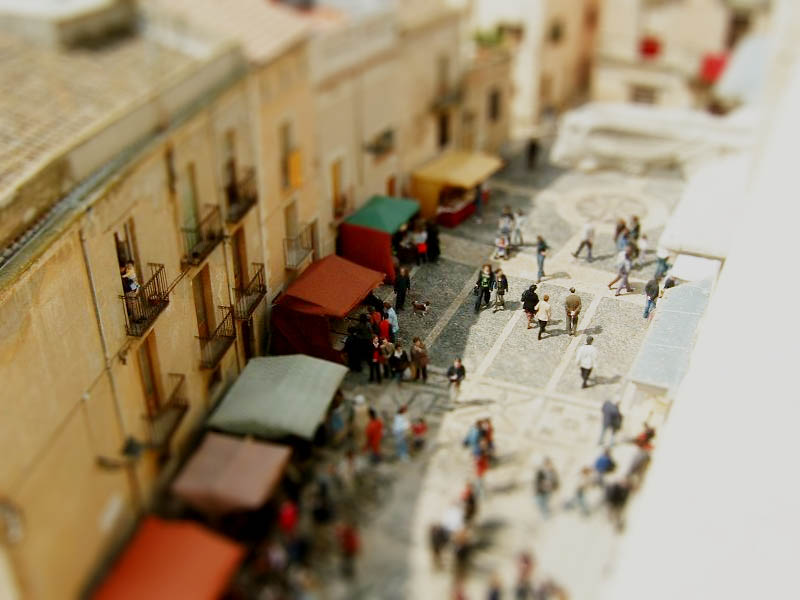}
  \caption{Focus matters! This photo is of a real scene, but the depth of field was reduced by a tilt-shift effect to make it look like a model \citep{Held:2009} (photo by oseillo).}
  \label{fig:focusmatters}
\end{figure}

Although it is usually forgotten in the list of depth cues, we should also add \emph{depth of field}, or retinal image blur \citep{Watt:2005,Hoffman:2008} (it is different from the accommodation cue cited before, which refers to the accommodation \emph{distance} only, not to the deph of field), the importance of which is well illustrated by Fig.\ref{fig:focusmatters}. The depth of field of the Human eye is around 0.3 Diopters (D) in normal situations, although finer studies~\cite{Marcos:1999} claim that it also slightly depends on  parameters such as the pupil size, wavelength, and spectral composition. Diopters are inverse of meters: at a focus distance of 3m, a depth of field of $\pm$ 0.3D, means that the in-focus range is from \(1/(\frac{1}{3}+0.3) \approx 1.6\textrm{m}\) to  \(1/(\frac{1}{3}-0.3) = 30\textrm{m}\), whereas at a focus distance of 30cm, the in-focus depth range is only from 27.5cm to 33cm (it is easy to understand from this formula why we prefer using diopters rather than a distance range to measure the depth of field: diopters are independent of the focus distance, and can easily be converted to a distance range). This explains why the photograph in Fig. \ref{fig:focusmatters} looks like a model rather than an actual-size scene \citep{Held:2010}: the in-focus parts of the scene seem to be only about 30cm away from the spectator. The depth of field range is not much affected by age, so this depth cue may be learned from observations over a long period, whereas the accommodation range goes from 12D for children, to 8D for young adults, to... below 1D for presbyopes.

\subsection{Stereoscopy and Stereopsis}
\label{sec:stereoscopy}

Stereoscopy\index{Stereoscopy}, i.e. the fact that we are looking at a scene using our two eyes, brings two additional physiological cues~ \cite{Lipton:1982}:
\begin{itemize}
\item \emph{Vergence} (the angle between the line-of-sight of both eyes);
\item \emph{Disparity} (the positional difference between the two retinal images of a scene point, which is non-zero for objects behind or in front of the convergence point).
\end{itemize}

These cues are used by the perception process called stereopsis\index{Stereopsis}, which gives a sensation of depth from two different viewpoints, mainly from the horizontal disparity.

Although stereoscopy and motion parallax are very powerful 3D depth cues, it should be noted that human observers asked to make judgments about the \threeD{} metric structure of a scene from these cues are usually subject to large systematic errors~\citep{Todd:2003,Sun:2009}.

\subsection{Conflicting Cues}
\label{sec:conflicting-cues}

All these cues (the 8 monoscopic cues and stereopsis) may be conflicting\index{Depth cues!conflicting}, i.e. giving opposite indications on the scene geometry. Many optical illusions make heavy use of these conflicting cues, i.e. when an object seems smaller because lines in the image suggest a vanishing point. Two famous examples are the Ames room and the pseudoscope.

\begin{figure}
  \centering
  \includegraphics[width=6cm]{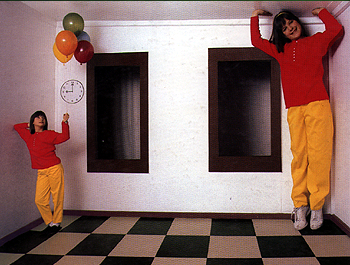}\includegraphics[width=5cm]{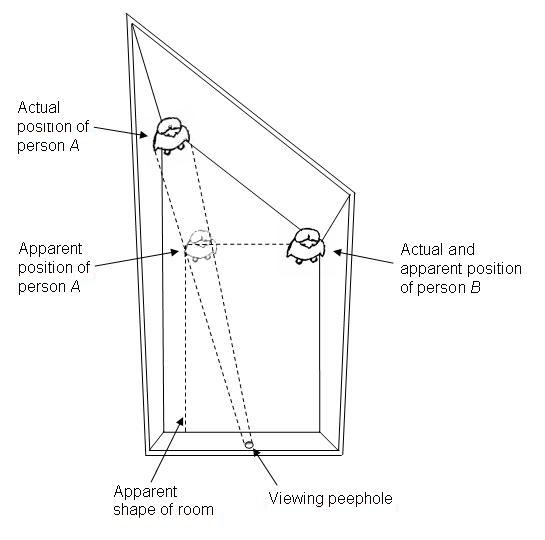}
  \caption{Ames room: an example of conflicting perspective and relative size cues.}
  \label{fig:amesroom}
\end{figure}

The Ames room (invented by Adelbert Ames in 1934) is an example where monocular cues are conflicting. Ames room is contained in a large box, and the spectator can look at it though a single viewpoint, which is a hole in one of the walls. From this viewpoint, this room seems to be cubic-shaped because converging lines in the scene suggest the three standard lines directions (the vertical, and two orthogonal horizontal directions), but it is really trapezoidal (fig.~\ref{fig:amesroom}). Perspective cues are influenced by prior knowledge of what a room should look like, so that persons standing in each far corner of the room will appear to be either very small or very big. The room itself can be seen from a peep hole at the front, thus forbidding binocular vision. In this precise case, binocular vision would easily disambiguate the conflicting cues by concluding that the room is not cubic. The Ames room was used in movies such as \textit{Eternal Sunshine of the Spotless Minds} by Michel Gondry or the \textit{Lord of the Rings} trilogy.

Another example of conflicting cues, which is more related to stereoscopic cinema, is illustrated by the pseudoscope. The pseudoscope (invented by Charles Wheatstone) is an binocular device which switches the viewpoints from the left and right eyes, so that all stereoscopic cues are reversed, but the monoscopic cues still remain and usually dominate the stereoscopic cues. The viewer still has the impression of ``seeing in \threeD{}'', and the closer objects in the scene actually seem bigger than they are, because the binocular disparity indicates that these big objects (in the image) are far away. This situation happened quite often during the projection of stereoscopic movies in the past~\cite{Zone:2005}, where the filters in front of the projectors or the film reels were accidentally reversed, but the audience usually did not notice what was wrong, and still had the impression of having seen a \threeD{} movie, though they thought it probably was a bad one because of the resulting headache. 

Conflicting perspective and stereoscopic cues were actually used heavily by Pete Kozachik, Director of Photography on Henry Selick's stereoscopic film Coraline, to give the audience different sensations~\cite{Kozachik:2009,Bordwell:2009}: ``Henry wanted to create a sense of confinement to suggest Coraline's feelings of loneliness and boredom in her new home. His idea had interiors built with a strong forced perspective and shot in \threeD{} to give conflicting cues on how deep the rooms really were. Later, we see establishing shots of the more appealing Other World rooms shot from the same position but built with normal perspective. The compositions match in \twoD{}, but the \threeD{} depth cues evoke a different feel for each room.''

\subsection{Inconsistent Cues}
\label{sec:non-consistent-cues}

Inconsistent cues\index{Depth cues!inconsistent} are usually less disturbing for the spectator than conflicting cues. They are defined as cues that indicate different amounts of depth in the same direction. They have been used for ages in bas-relief, where the lighting cue enhances the depth perceived by the binocular system, and as a matter of fact bas-relief is usually better appreciated from a far distance, where the stereoscopic cues have less importance.

An effect that is often observed when looking at stereoscopic photographs is called the \emph{cardboard effect}\index{Cardboard effect} \cite{Yamanoue:2006,Masaoka:2006}: some depth is clearly perceived, but the amount of depth is too small with respect to the expected depth from the image size of the objects, resulting in objects appearing as flat, or drawn on cardboard of billboards. We will explain later how to predict this effect, and most importantly how to avoid it.

Another well-known stereoscopic effect is called the \emph{puppet-theater effect}\index{Puppet-theater effect} (also called \emph{pinching}\index{Pinching effect|see{Puppet-theater effect}}): background objects do not appear as small as expected, so that foreground objects appear proportionately smaller.

These inconsistent cues can easily be avoided if there is total control on the shooting geometry, including camera placement. If there are some unavoidable constraints on the shooting geometry, we will explain in Sec.~\ref{cha:chang-shoot-geom} how \emph{some} of these inconsistent cues related to stereoscopy can be corrected in post-production (Sec.~\ref{cha:chang-shoot-geom}).

\subsection{Sources of Visual Fatigue}
\label{cha:sour-visu-fatig}

Visual fatigue is probably the most important point to be considered in stereoscopic cinema. Stereoscopic movies in the past often resulted in a bad viewing experience, and this reduced a lot the acceptance of stereoscopic cinema by a large public. \citet{Ukai:2007} produced a reference study on visual fatigue caused by viewing stereoscopic films, and is a good introduction to this field.

The symptoms of visual fatigue may be conscious (headache, tiredness, soreness of the eyes) or unconscious (perturbation of the oculo-motor system). It should actually be considered as a public health concern~\cite{Wann:1997}, just as the critical fusion frequency on CRT screens 50 years ago, as it may actually lead to difficulties in judging distances (which is very important in such tasks as driving). \citet[Sec. 6]{Ukai:2007} even report the case of an infant whose oculo-motor system was permanently disturbed by viewing a stereoscopic movie. Although the long-term effects of viewing stereoscopic cinema were not studied due to the fact that this medium is not yet widespread, many studies exist on the effects on health of using virtual reality displays \cite{Wann:1995,Wann:1997,Lambooij:2007a}. Virtual reality displays are widely used in the industry (from desktop displays to immersive displays), and are sometimes used daily by people working in industrial design, data visualization, or simulation.

The sources of visual fatigue that are specific to stereoscopic motion pictures are mainly due to binocular asymmetry, i.e. photometric or geometric differences between the left and right retinal images. \citet{Kooi:2004} experimentally measured thresholds on the various assymetries that will lead to visual incomfort (incomfort is the lowest grade of conscious visual fatigue). For example, they measured, in agreement with Pastoor's rule of thumb~\cite{Pastoor:1992} that a 35 arcmin horizontal disparity range is quite acceptable for binocular perception of \threeD{} and 70 arcmin disparity is too much to be viewed. They also found out that the human visual system is most sensitive to vertical binocular disparities. The various quantitative binocular thresholds computed from their experiments can be found in \citet[Table 4]{Kooi:2004}.

\begin{figure}
  \centering
  \includegraphics[width=0.3\columnwidth]{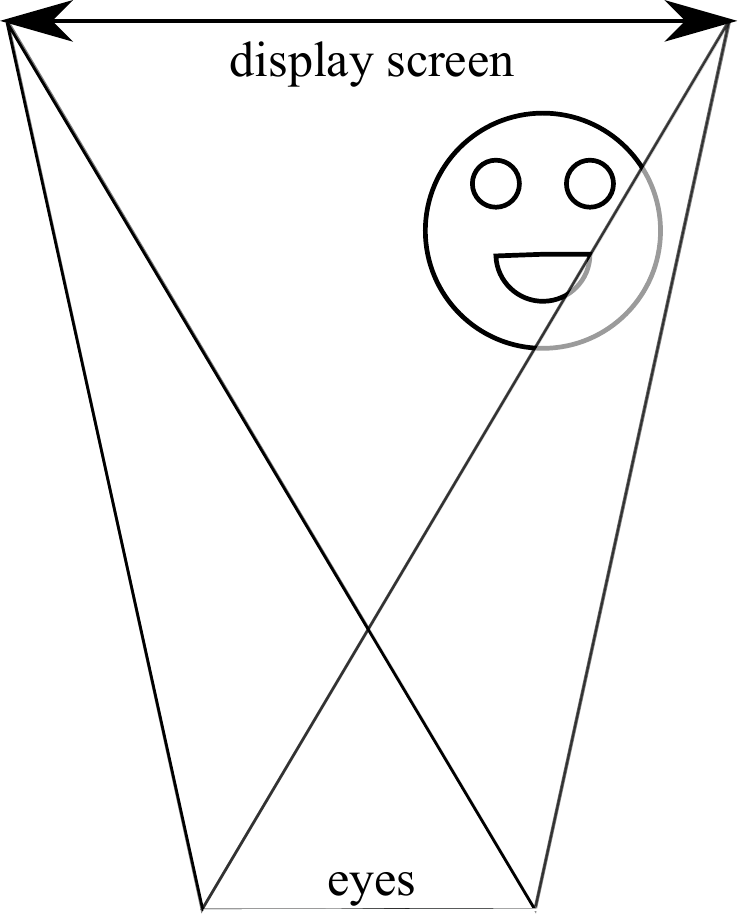} \includegraphics[width=0.3\columnwidth]{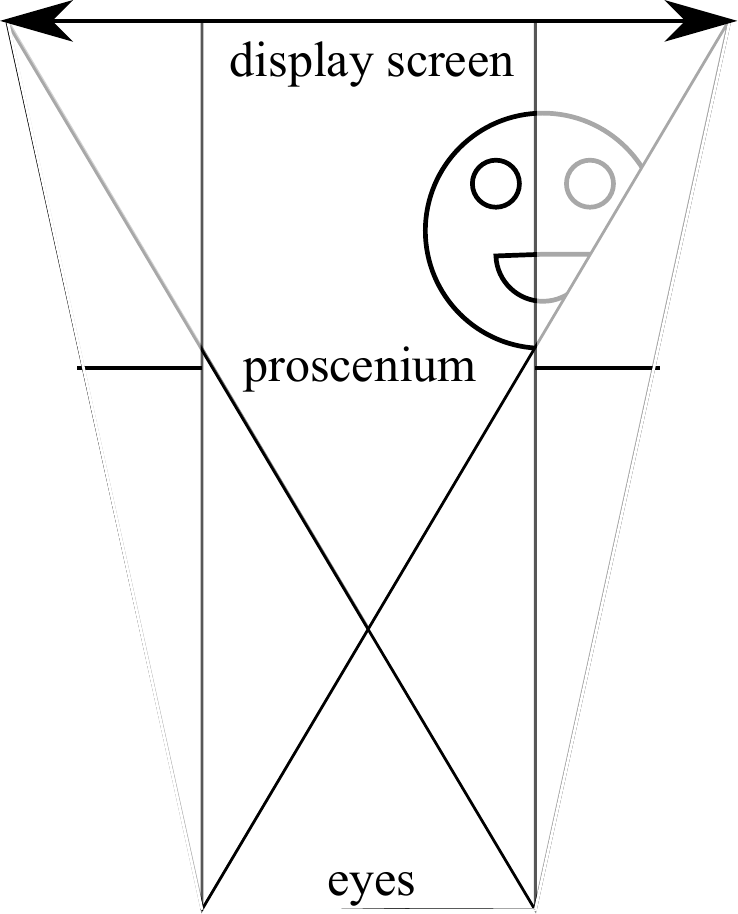}
  \caption{Breaking the proscenium rule: (left) part of the object in front of the proscenium arch is not visible in one eye, which breaks the proscenium rule; (right) masking part of the image in each eye moves the proscenium closer than the object, and the proscenium rule is re-established.}
  \label{fig:proscenium}
\end{figure}

The main sources of visual fatigue can be listed as:
\begin{itemize}
\item \emph{Crosstalk}\index{Stereoscopic cinema!crosstalk} (sometimes called \emph{crossover} or \emph{ghosting}), which is usually due to a stereoscopic viewing system with a single screen: a small fraction of the intensity from the left image can be seen in the right eye, and vice-versa. The typical values for crosstalk~\cite{Kooi:2004} are 0.1-0.3\% with polarization-based systems, and 4-10\% with LCD shutter glasses. Preprocessing can be applied to the images before displaying in order to reduce crosstalk by subtracting a fraction of the left image from the right image~\cite{Mendiburu:2009}  -- a process sometimes called \emph{ghost-busting}.

\item \emph{Breaking the proscenium rule}\index{Proscenium arch} (or breaking the stereoscopic window) happens when there are interposition errors between the stereoscopic imagery and the edges of the display (see Fig.~\ref{fig:proscenium} and \citet[Chap. 5]{Mendiburu:2009}).  A simple way to correct this is to move the proscenium closer to the spectator by adding borders to the image, but this is not always as easy as it seems (see Sec.~\ref{sec:break-prosc-rule}).

\item \emph{Horizontal disparity limits} are the minimum and maximum disparity values that can be accepted without producing visual fatigue. An obvious bound for disparity is that the eyes should not diverge. Another limit concerns the range of acceptable disparities within a stereoscopic scene that can be fused simultaneously by the human visual system.

\item \emph{Vertical disparity} causes torsion motion of the ocular globes, and is only tolerable for short time intervals. Our oculomotor system learned the epipolar geometry of our eyes over a lifetime of real-world experience, and any deviation from the learned motion causes strain.

\item \emph{Vergence-accomodation conflicts} occur when the focus distance of the eyes is not consistent with their vergence angle. It happens quite often when viewing stereoscopic cinema, since the display usually consists of a planar surface placed at a fixed distance. Strictly speaking, any \threeD{} point that is not in the convergence plane will have an accomodation distance, which is exactly the screen distance, different from the vergence distance. However, this constraint can be somewhat relaxed by using the depth of field of the visual system, as will be seen later.

\end{itemize}

\begin{figure}
  \centering
  \includegraphics[scale=0.9]{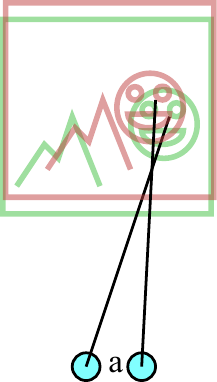}
  \includegraphics[scale=0.9]{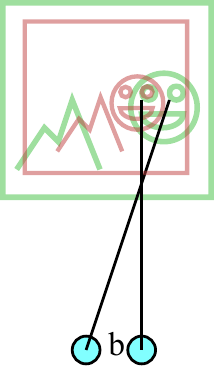}
  \includegraphics[scale=0.9]{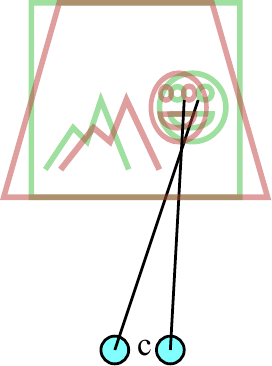}
  \includegraphics[scale=0.9]{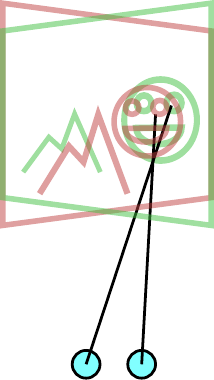}
  \includegraphics[scale=0.9]{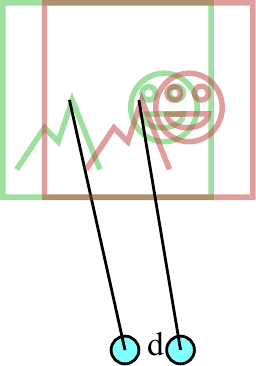}

  \caption{A few examples of geometric asymmetries: (a) Vertical shift, (b) Size or magnification difference, (c) Distortion difference, (d) Keystone distortion due to toed-in cameras, (e) Horizontal shift - leading to eye divergence in this case (adapted from \citet{Ukai:2007}).}
  \label{fig:ukai-howarth-geometry}
\end{figure}

Geometric asymmetries come very often either from a misalignment or from a difference between the optics in the camera system or in the projection system, as seen in Fig.~\ref{fig:ukai-howarth-geometry}.
In the following, we will only discuss \emph{horizontal disparity limits}, \emph{vertical disparity}, and \emph{vergence-accommodation conflicts}, since the other sources of visual fatigue are easier to deal with.

\subsubsection{Horizontal Disparity Limits}
\label{sec:horiz-disp-limits}

The most simple and obvious disparity limit is eye divergence. In their early work on stereoscopic cinema, the Spottiswoodes said: ``It is found that divergence is likely to cause eyestrain, and therefore screen parallaxes in excess of the eye separation should be avoided''. But they also went on to say, in listing future development requirements, that ``Much experimental work must be carried out to determine limiting values of divergence at different viewing distances which are acceptable without eyestrain''.

These limiting values are the maximum disparities acceptable around the convergence point, usually expressed as angular values, such that the binocular fusion of the \threeD{} scene is performed without any form of eyestrain or visual fatigue. Many publications dealt with the subject of finding the horizontal disparity limits~\citep{Yeh:1990,Jones:2004,Pastoor:1992}.

The horizontal disparity limits are actually closely related to the depth of field, as noted by \citet{Lambooij:2007a}: ``An accepted limit for DOF in optical power for a 3 mm pupil diameter (common under normal daylight conditions) and the eyes focusing at infinity, is one-third of a diopter. With respect to the revisited Panum's fusion area\footnote{In the human visual system, the space around the current fixation point which can be fused is called Panum's area or fusion area. It is usually measured in minutes of arc (arcmin).}, disparities beyond one degree (a conservative application of the 60 to 70 arcmin recommendation), are assumed to cause visual discomfort, which actually results from the human eye's aperture and depth of field. Though this nowadays serves as a rule-of-thumb, it is acknowledged as a limit, because lower recommendations have been reported as well. If both the limits of disparity and DOF are calculated in distances, they show very high resemblance.''.
\citet{Yano:2004} also showed that images containing disparities beyond the depth of field ($\pm$ 0.2D depth of field, which means $\pm 0.82^\circ$ in disparity) cause visual fatigue.

\subsubsection{Vertical Disparity}

Let us suppose that the line joining both eyes is horizontal, and that the stereoscopic display screen is vertical and parallel to this line. The images of any \threeD{} point projected onto the display screen using each eye optical center as the centers of projection are two points which are aligned horizontally, i.e. have no vertical disparity. Thus all the scene points that are displayed on the screen should have no vertical disparity. Vertical disparity (see Fig.~\ref{fig:ukai-howarth-geometry} for some examples) may come from a misalignment of the cameras or of the display devices, from a focal length difference between the optics of the cameras, from keystone distortion, due to a toed-in camera configuration, or from nonlinear (e.g. radial) distortions.

However, it is to be noted that vertical disparities exist in the visual system: remember that the eye is not a linear perspective camera, but a spheric sensor, so that an object which is not in the median plane between both eyes will be closer to one eye than to the other, and thus its image will be bigger in one eye than in the other (a spheric sensor basically measures angles). The size ratio between the two images is called the vertical size ratio, or VSR. VSR is naturally present when rectified images (i.e. with no vertical disparity) are projected on a flat display screen: a vertical rod, though it's displayed with the same size on the left and right images on the flat display, subtends a larger angle in the nearest eye.

Psychophysical experiments showed that vertical disparity gradients have a strong influence on the perception of stereoscopic shape, depth and size \citep{Ogle:1938,Rogers:1993,Allison:2003}. For example, the so-called induced-size effect \citep{Ogle:1938} is caused by a vertical gradient of vertical disparity (vertical-size parallax transformation) between the half images of an isolated surface, which creates an impression of a surface slanted in depth. \citet{Ogle:1938,Ogle:1964} called it the induced-size effect because it is as though the vertical magnification of the image in one eye induces an equivalent horizontal magnification of the image in the other eye\footnote{However, \citet{Read:2006} show that non-zero vertical disparity sensors are in fact not necessary to explain the induced size effect.}.

\citet{Allison:2004} also notes that vertical disparities can be used to fool the visual system: ``Images on the retinae of a fronto-parallel plane placed at some distance actually have some keystone distortion, which may be used as a depth cue. Displaying keystone-distorted images on that fronto-parallel screen actually exaggerates that keystone distortion when the viewer is focusing on the center of the screen, and would thus giving a cue that the surface is nearer than the physical screen.''.

By displaying keystoned images, the VSR is distorted in a complicated way which may be inconsistent with the horizontal disparities. Besides, that distortion depends on the viewer position with respect to the screen. When the images displayed on the screen are rectified, although depth perception may be distorted depending on the viewer position, horizontal and vertical disparities will always be consistent, as long as the viewer's interocular (the line joining the two optical centers) is kept parallel to the screen and horizontal (this viewing position may be hard to obtain in the side rows of wide movie theaters).

\citet{Woods:1993} discuss sources of distortion in stereo camera arrangements as well as the human factors considerations required when creating stereo images. These experiments show that there is a limit in the screen disparity which it is comfortable to show on stereoscopic displays. A limit of 10 mm screen disparity on a 16'' display at a viewing distance of 800 mm was found to be the maximum that all 10 subjects of the experiment could view. Their main recommendation is to use a parallel camera configuration in preference to converged cameras, in order to avoid keystone distortion and depth plane curvature.

\textbf{Is Vertical Disparity Really a Source of Visual Fatigue?}
From the fact that vertical disparities are actually a depth cue, there has been a debate on whether vertical disparities can be a source of visual fatigue, since they are naturally present in retinal images. \citet{Stelmach:2003} and \cite{Speranza:2002} claim that keystone and depth plane curvature cause minimal discomfort: images plane shift (equivalent to rectified images) and toed-in cameras are equally comfortable in their opinion.  However, we must distinguish between visual discomfort, which is conscious, and visual fatigue, where the viewer may not be conscious of the problem during the experiment, but headache, eyestrain, or long-term effects can happen.

Even in the case where the vertical disparities are not due to a uniform transform of the images, such as a rotation, scaling or homography, \citet{Stevenson:1997} demonstrated that human stereo matching does not actually follow the epipolar lines, and human subjects can still make accurate near/far depth discrimination when the vertical disparity range is as high as 45 arcmin.

\citet{Allison:2007} concludes that, although keystone-distorted images coming from a toed-in camera configuration can be displayed with their vertical disparities without discomfort, the images should preferably be rectified, because the additional depth cues caused by keystone distortion perturb the actual depth perception process.

\subsubsection{Vergence-Accommodation Conflicts}
\label{sec:verg-accom-confl}

When looking at a real \threeD{} scene, the distance of accomodation (i.e. the focus distance) is equal to the distance of convergence, which is the distance to the perceived object. This relation between these two oculo-motor functions, called Donder's line~\cite{Yano:2004}, is learned through the first years of life, and is used by the visual system to quickly focus-and-converge on objects surrounding us. The relation between vergence and accomodation does not have to follow exactly Donder's line: there is an area around it where vergence and accomodation agree, which is called Percival's zone of comfort~\cite{Hoffman:2008,Yano:2004}.

\begin{figure}
  \centering
  \includegraphics[scale=0.6]{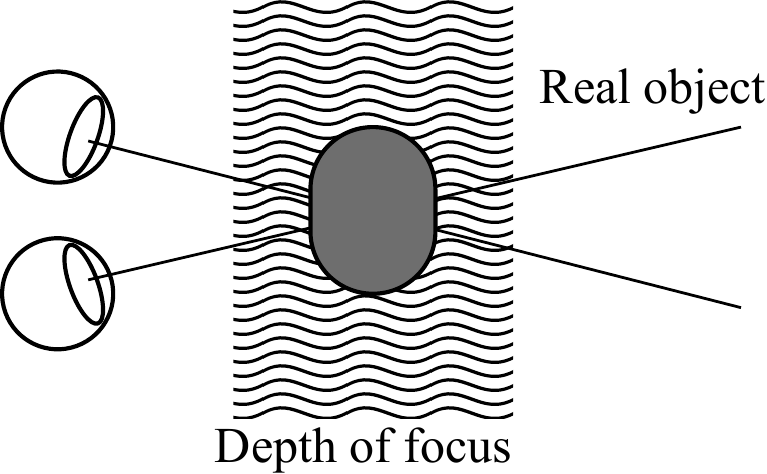} \hfill \includegraphics[scale=0.6]{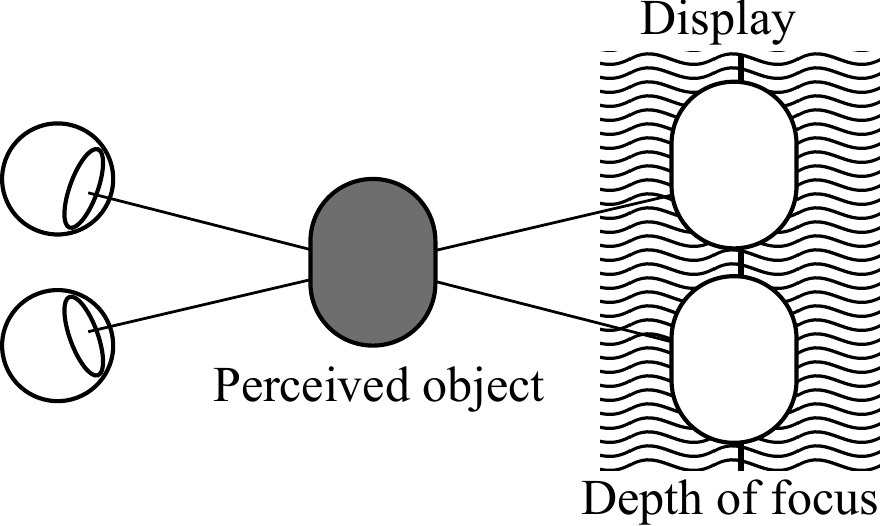}
  \caption{Vergence and accomodation: they are consistent when viewing a real object (left), but may be conflicting when viewing a stereoscopic display, since the perceived object may not lie within the depth of field range (right). Adapted from \citet{Emoto:2005}.}
\label{fig:dof}
\end{figure}

When viewing a stereoscopic movie, the distance of accomodation differs from the distance of convergence, which is the distance to the perceived object (Fig.~\ref{fig:dof}). This discrepancy causes a perturbation of the oculo-motor system~\cite{Yano:2004,Emoto:2005}, which causes visual fatigue, and may even damage the visual acuity, which is reported by \citet{Ukai:2007} to have plasticity until the age of 8 or later. This problem has been largely studied for virtual reality (VR) displays, and lately for \threeD{} television (3DTV), but has been largely overlooked in movie theater situations. In fact, many stereoscopic movies, especially IMAX-3D movies, make heavy use of spectacular effects by presenting perceived objects which are very close to the spectator, when the screen distance is about 20m.

\citet{Wann:1997} cite it as one of the main sources of stress in VR displays. They observed that, in a situation where other sources of visual discomfort were eliminated, prolonged use of a stereoscopic display caused short-term modifications in the normal accommodation-vergence relationship.

\citet{Hoffman:2008} designed a special \threeD{} display where vergence and focus cues are consistent \citep{Akeley:2004}. The display is designed so that its depth of field approximately corresponds to the Human depth of field. By using this display, they showed that ``when focus cues are correct or nearly correct, (1) the time required to identify a stereoscopic stimulus is reduced, (2) stereo-acuity in a time-limited task is increased, (3) distortions in perceived depth are reduced, and (4) viewer fatigue and discomfort are reduced.'' However, this screen is merely experimental, and it is impractical for home or movie theater use.

The depth of field may be converted to disparities in degrees using simple geometric reasoning. The depth of field of the human visual system is, depending on the authors, between $\pm$ 0.2D or $\pm 0.82^\circ$ \cite{Yano:2004} and $\pm$ 0.3D \cite{Campbell:1957}. The limit to binocular fusion is from 2 to 3 degrees at the front or back of the stereoscopic display, and Percival's zone of comfort is about one third of this, i.e. 0.67 to 1 degree.  We note that the in-focus range almost corresponds to Percival's zone of comfort for binocular fusion, which probably comes from the fact that the visual system only learned to fuse non-blurred objects within the in-focus range.

Let us take for example a conservative value of $\pm$ 0.2D for the depth of field. For a movie theater screen placed at 16m, the in-focus range is from \(1/(1/16+0.2) \approx 3.8\mathrm{m}\) to infinity, whereas for a 3DTV screen placed at 3.5m, it is from \(1/(1/3.5+0.2) \approx 2\mathrm{m}\) to \(1/(1/3.5+0.2) \approx 11.7\mathrm{m}\). As we will see later, this means that the camera focus range should theoretically be different when shooting movies for a movie theater or a \threeD{} television.

\section{Picking the Right Shooting Geometry}
\label{cha:keep-prop-pick}

\subsection{The Spottiswoode Point of View}
\label{sec:spott-point-view}

\citet{Spottiswoode:1952} wrote the first essay on the perceived geometry in stereoscopic cinema. They devised how depth is distorted by ``stereoscopic transmission'' (i.e. recording and reproduction of a stereoscopic movie), and how to achieve ``continuity in space'', or making sure that there is a smooth transition in depth when switching from one stereoscopic shot to another.

They did a strong criticism of the ``human vision'' systems (i.e. systems that were trying to mimic the human eyes, with a 6.5cm interocular, and a $0.3^{\circ}$ convergence). They claimed that all stereoscopic parameters can be and should be adapted, either at shooting time or as post-corrections, depending on screen size, to get the desirable effects (depth magnification or reduction, and continuity in space). According to them, the main stereoscopic parameter is the \emph{nearness factor} $N$, defined as the ratio between the viewing distance from screen and the distance to the fused image (with our notations, \(N=H'/Z'\)): ``for any pair of optical image points, the ratio of the spectator's viewing distance from the screen to his distance to the fused image point is a constant, no matter whereabouts in the theater he may be sitting''. Continuity in space is achieved by slowly shifting over a few seconds the images in the horizontal direction before and after the cuts. The spectator does not notice that the images are slowly shifting: the vergence angle of the eyes is adapted by the human visual system, and depth perception is almost the same. Due to the technical tools available at this time, this is the only post-correction method they propose.

They list three classes of stereoscopic transmission:
\begin{itemize}
\item \emph{ortho-infinite}, where infinity points are correctly represented at infinity,
\item \emph{hyper-infinite}, where objects short of infinity are represented at infinity (which means that infinity points cause divergence),
\item \emph{hypo-infinite}, where objects at infinity are represented closer than infinity (which causes the cardboard effect\index{Cardboard effect}, see Sec.~\ref{sec:non-consistent-cues}).
\end{itemize}

In order to help the stereographer picking up the right shooting parameters, they invented a calculator, called the \emph{Stereomeasure}, which computes the relation between the various parameters of \threeD{} recording and reproduction. They claim that a stereoscopic movie has to be made for given projection conditions (screen size and distance to screen), which is a fact sometimes overlooked either by stereographers, or by the 3DTV industry (however, as discussed in Sec.~\ref{cha:chang-shoot-geom}, the stereoscopic movie may be adapted to other screen sizes and distances).

In the Spottiswoode setup, the standard distance from spectator to screen should be from $2W'$ to $2.5W'$ ($W'$ is the screen width). They place the proscenium arch\index{Proscenium arch} at $N=2$ (half distance from screen), and almost everything in the scene happens between $N=0$ (infinity) and $N=2$ (half distance) ($N=1$ is the screen plane).

Their work has been strongly criticized by many stereographers, sometimes with wrong arguments~\cite{Lipton:1982,Smith:1983}, but the main problem is probably that the artists do not want to be constrained by mathematics when they are creating: cinematography has always been an art of freedom, with a few rules of thumb that could always be ignored. But the reality is here: the constraints on stereoscopic cinema are much stronger than on \twoD{} cinema, and bypassing the rules results in bad movies causing eyestrain or headache to the spectator. A bad stereoscopic movie can be a very good \twoD{} movie, but adding the stereoscopic dimension will always modify the perceived quality of the movie, either by adding a feeling of ``being there'', or by obfuscating the intrinsic qualities of the movie with ill-managed stereoscopy.

There are some problems with the Spottiswoode's theorisation of stereoscopic transmission, though, and this section will try to shed some light on some of these:
\begin{itemize}
\item the parametrization by the nearness factor hides the fact that strong nonlinear depth and size distortions may occur in some cases, especially on far points;
\item divergence at infinity will happen quite often when images are shifted in order to achieve continuity in space;
\item shifting the images may break the vergence-accomodation constraints, in particular Percival's zone of comfort, and cause visual fatigue.
\end{itemize}

\citet{Woods:1993} extended this work and also computed spatial distortion of the perceived geometry. Although their study is more focused on determining horizontal and vertical disparity limits,
they also studied  depth plane curvature effects, where a fronto-parallel plane appears to be curved. This situation arises when non-rectified images are used and the camera configuration is toed-in (i.e. with a non-zero vergence angle).

More recently, \citet{Masaoka:2006} from the NHK labs also did a similar study, and presented a software tool which is able to predict spatial distortions that happen when using given shooting and viewing parameters.

\subsection{Shooting and Viewing Geometries}
\label{sec:shoot-view-geom}

As was shown by \citet{Spottiswoode:1952} in the early days of stereoscopic cinema, projecting a stereoscopic movie on different screen sizes and distances will produce different perceptions of depth.

One obvious solution was adopted by the large format stereoscopic cinema (IMAX \threeD{}): if the film is shot with parallel cameras (i.e. vergence is zero), and is projected with parallel projectors that have a human-like interocular (usually 6cm for IMAX-3D), then infinity points will always be perceived exactly at infinity, and divergence will never occur~\cite{Zone:2005}. Large-format stereoscopic cinema has less constraints than stereoscopic cinema targeted at standard movie theaters, since the screen is practically borderless, and the audience is located near the center of the hemispherical screen. The camera interocular is usually close to the human interocular, but it may be played with easily, depending on the scene to be shot (as in \emph{Bugs! \threeD{}} by Phil Streather, where hypostereo or gigantism\index{Gigantism} is heavily used). If the camera interocular is the same as the human interocular, then everything in the scene will appear at the same depth and size as if seen with normal vision, which is a very pleasant experience.

However, shooting parallel is not always advisable or even possible for standard stereoscopic movies. Close scenes for example, where the subject is closer than the movie theater screen, require camera convergence, or the film subject will appear too close to the spectator and will break the proscenium rule most of the time. But using camera convergence has many disadvantages, and we will see that it may cause heavy distortions on the \threeD{} scene.

\begin{figure}[htbp]
  \centering
  \begin{tabular}{cc}
  \imagecenter{\includegraphics[width=0.35\columnwidth]{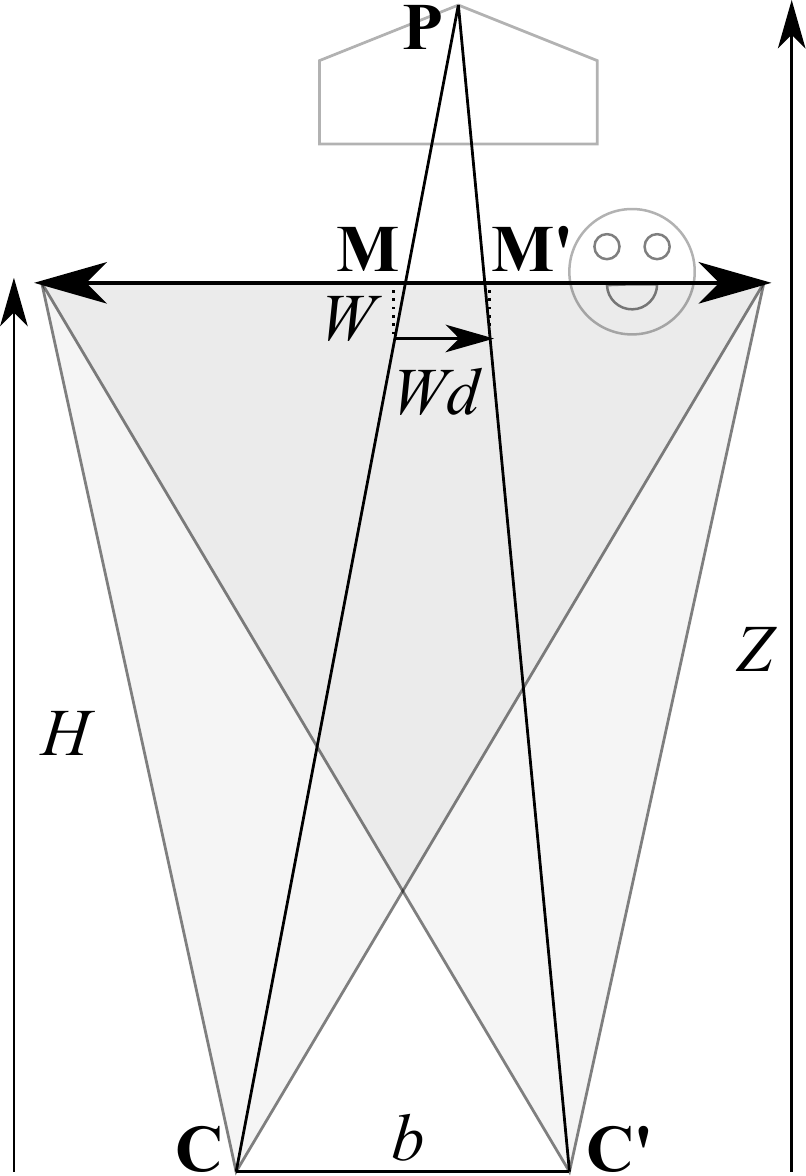}}\quad\quad &
  \begin{tabular}{|c|c|c|}
    \hline
    Symbol & Camera & Display \\
    \hline \hline
    $\mathbf{C}$, $\mathbf{C'}$ & camera optical center & eye optical center \\
    \hline
    $\mathbf{P}$ & physical scene point & perceived \threeD{} point \\
    \hline
    $\mathbf{M}$, $\mathbf{M'}$ & image points of $\mathbf{P}$ & screen points \\
    \hline
    $b$ & camera interocular & eye interocular \\
    \hline
    $H$ & convergence distance & screen distance \\
    \hline
    $W$ & width of convergence plane & screen size \\
    \hline
    $Z$ & real depth & perceived depth \\
    \hline
    $d$ & \multicolumn{2}{c|}{disparity (as a fraction of $W$)} \\
    \hline
  \end{tabular} 
  \end{tabular}
  \caption{Shooting and viewing geometries can be described using the same small 
set of parameters.}
  \label{fig:geometry}
\end{figure}

Let us study the distortions caused by given shooting and viewing geometries. The geometric parameters we use (Fig.~\ref{fig:geometry}) are very simple but describe fully the stereoscopic setup. Compared to camera-based parameters, as used by \citet{Yamanoue:2006}, we can easily attach a simple meaning to each parameter and understand its effect on space distortions. We assume that the stereoscopic movie is rectified and thus contains no vertical disparity (see Sec.~\ref{cha:elim-vert-disp}), so that the convergence plane, where the disparity is zero, is vertical and parallel to the line joining the optical centers of the cameras.

The \threeD{} distortions in the perceived scene essentially come from different scene magnifications in the fronto-parallel (or width and height) directions, and in the depth direction. \citet{Spottiswoode:1952} defined the \emph{shape ratio} as the ratio between depth magnification and width magnification, \citet{Yamanoue:2006} call it $E_p$ or \emph{depth reduction}, and \citet{Mendiburu:2009} uses the term \emph{roundness factor}\index{Roundness factor}. We will use \emph{roundness factor} in the remaining of our study. A low roundness factor will result in the cardboard effect, and a rule of thumb used by stereographers is that it should never be below 0.2, or 20\%.

Let $b$, $W$, $H$, $Z$ be the stereoscopic camera parameters, and $b'$, $W'$, $H'$, $Z'$ be the viewing parameters, as described on Fig.~\ref{fig:geometry}. Let $d$ be the disparity in the images. The disparity on the display screen is \(d' = d + d_0\), taking into account an optional shift $d_0$ between the images (shifting can be done at the shooting stage, in post-production, or at the display stage).

Triangles $\mathbf{MPM'}$ and $\mathbf{CPC'}$ are homothetic, consequently:
\begin{equation}
  \label{eq:1}
  \frac{Z-H}{Z} = \frac{W}{b}d.
\end{equation}
It can easily be rewritten to get the image disparity $d$ as a function of the real depth $Z$ and vice-versa:
\begin{equation}
  \label{eq:2}
  d = \frac{b}{W}\frac{Z-H}{Z}, \text{ or } Z = \frac{H}{1 - \frac{W}{b}d},
\end{equation}
and the perceived depth $Z'$ as a function of the disparity $d'$:
\begin{equation}
  \label{eq:4}
  Z'=\frac{H'}{1-\frac{W'}{b'}(d+d_0)}.
\end{equation}
Finally, eliminating the disparity $d$ from both equations gives the relation between real depth $Z$ and perceived depth $Z'$:
\begin{equation}
  \label{eq:5}
  Z' = \frac{H'}{1 - \frac{W'}{b'} ( \frac{b}{W}\frac{Z-H}{Z}+d_0 ) } \quad\text{or}\quad
  Z = \frac{H}{1 - \frac{W}{b} ( \frac{b'}{W'}\frac{Z'-H'}{Z'}-d_0 ) }
\end{equation}

\subsection{Depth Distortions}
\label{sec:depth-distortions}

Let us now compute the depth distortion from the perceived depth. In the general case, points at infinity (\(Z \rightarrow +\infty\)) are perceived at:
\begin{equation}
  \label{eq:7}
  Z'=\frac{H'}{1-\frac{W'}{b'}(\frac{b}{W}+d_0)}.
\end{equation}

Eye divergence happens when $Z'$ becomes negative. The eyes diverge when looking at scene points at infinity (\(Z \rightarrow +\infty\)) if and only if:
\begin{equation}
  \label{eq:8}
  \frac{b'}{W'} < \frac{b}{W}+d_0.
\end{equation}
In this case, the real depth which is mapped to \(Z' = \infty\) can be computed from (\ref{eq:5}) as:
\begin{equation}
  \label{eq:12}
  Z = \frac{H}{1-\frac{W}{b}\left(\frac{b'}{W'}-d_0\right)},
\end{equation}
and any object at a depth beyond this one will cause divergence.

The relation between $Z$ and $Z'$ is nonlinear, except if \(\frac{W}{b} = \frac{W'}{b'}\) and \(d_0 = 0\), which we call the \emph{canonical setup}. In this case, the relation between $Z$ and $Z'$ simplifies to:
\begin{equation}
  \label{eq:6}
   Z' = Z\frac{H'}{H}.\qquad\text{(Canonical setup:}\quad \frac{W}{b} = \frac{W'}{b'}, d_0=0\text{)}
\end{equation}

Let us now study how depth distortion behaves when we are not in the canonical setup anymore. For our case study, we start from a purely canonical setup, with \(b=b'=6.5cm\), \(W=W'=10m\), \(H=H'=15m\) and \(d_0=0\). In the following charts, depth is measured from the convergence plane or the screen plane (depth increases away from the cameras/eyes).

\begin{figure}[hbtp]
  \centering
  \includegraphics[width=\columnwidth]{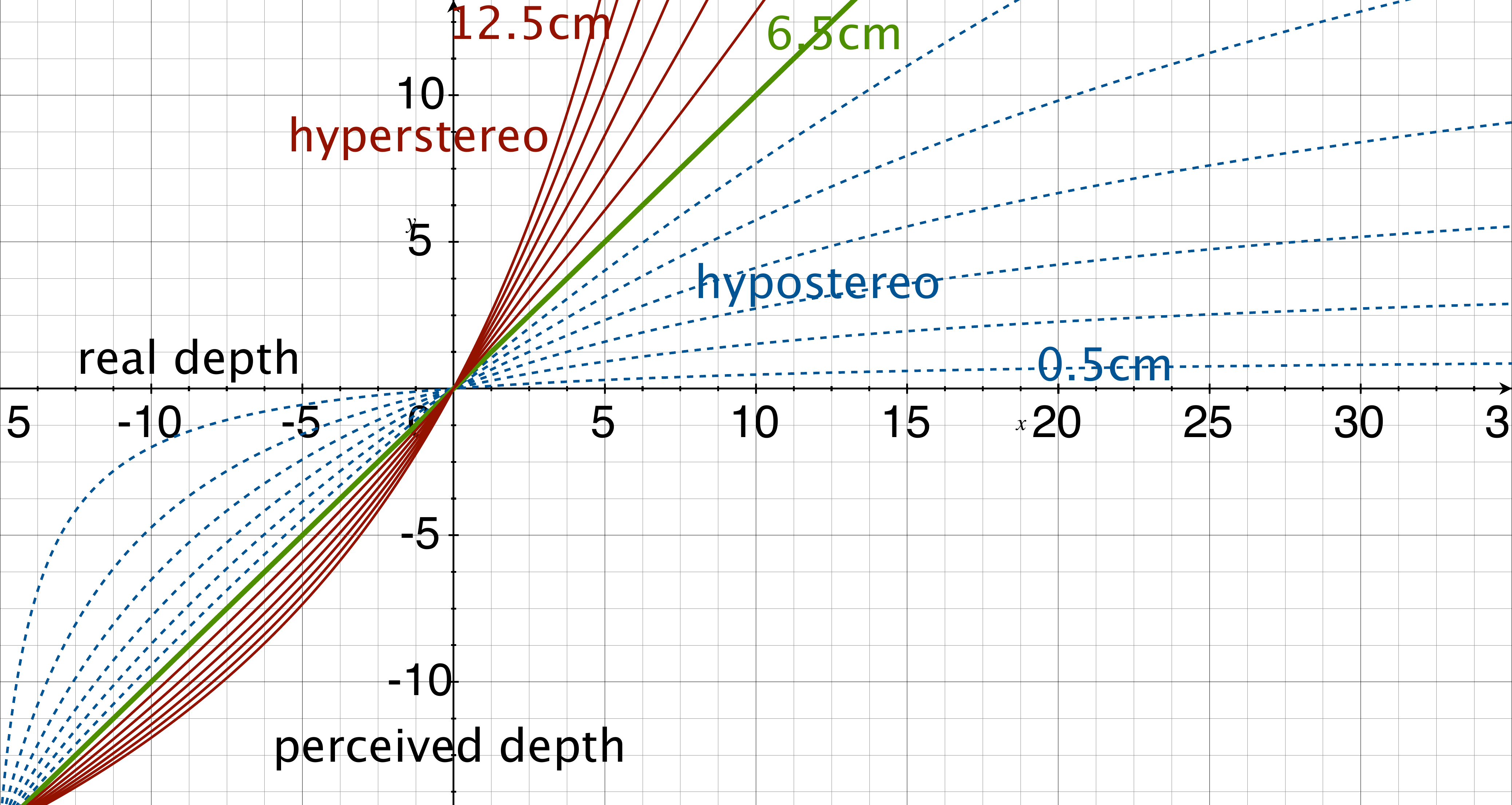}
  \caption{Perceived depth as a function of real depth for different values of the camera interocular $b$. This graph demonstrates the well-known hyperstereo and hypostereo phenomenon.}
  \label{fig:interocular}
\end{figure}

The first chart (Fig.~\ref{fig:interocular}) shows the effect of changing only the camera interocular $b$ (the distance to the convergence plane $H$ remains unchanged, which implies that the vergence angle changes accordingly). This shows the well-known effects called \emph{hyperstereo}\index{Hyperstereo} and \emph{hypostereo}\index{Hypostereo}: the roundness factor of in-screen object varies from high values (hyperstereo) to low values (hypostereo).

\begin{figure}[hbtp]
  \centering
    \includegraphics[width=\columnwidth]{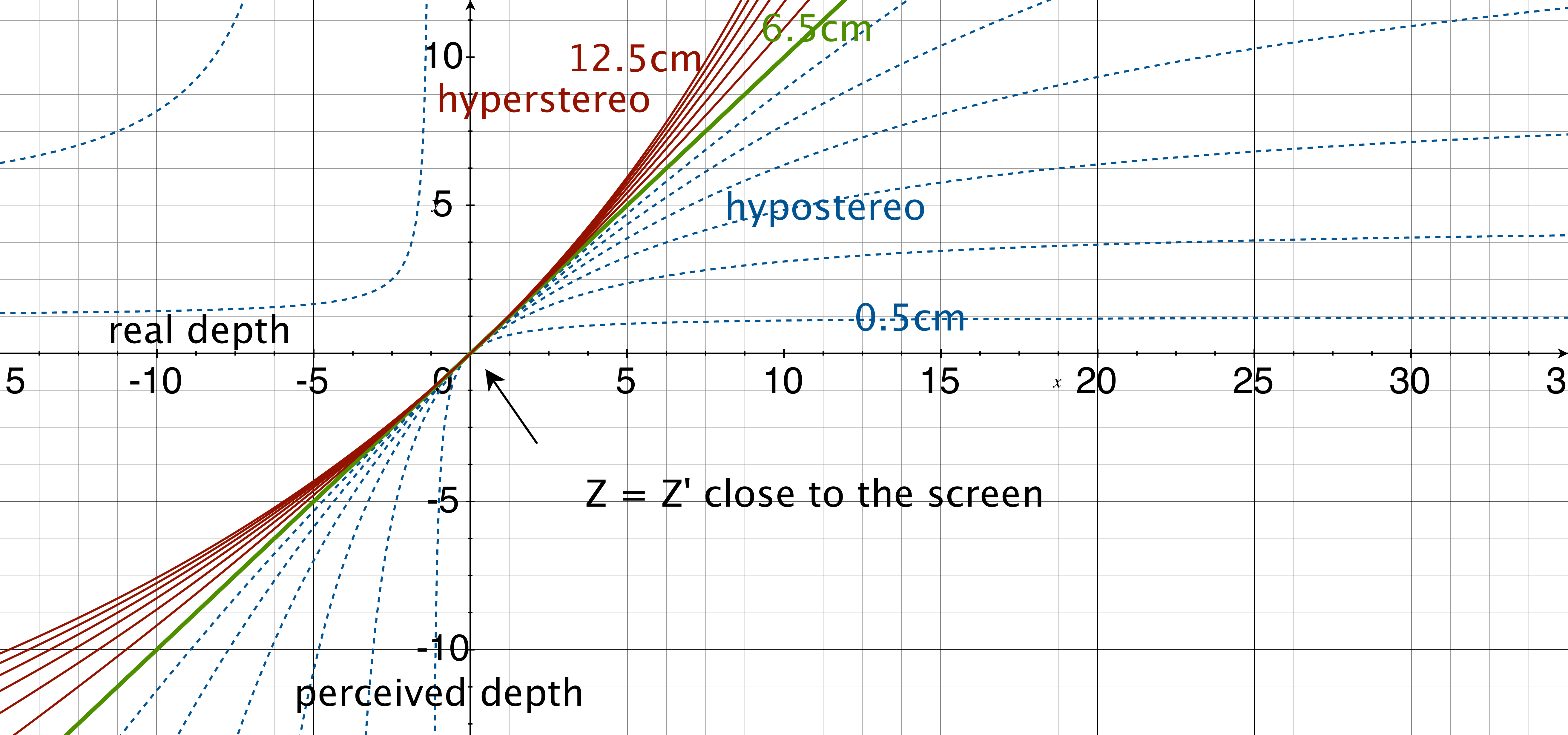}
  \caption{When the object moves away, but we keep the object image size constant by zooming in (i.e. $H$ varies but $W$ is constant), if we keep the camera baseline $b$ proportional to convergence distance $H$, the depth magnification close to the screen plane remains equal to 1. But be careful: divergence may happen!}
  \label{fig:zoomingin}
\end{figure}

Let us now suppose that the subject being filmed is moving away from the camera, and we want to keep its image size constant by adjusting the zoom and vergence accordingly, i.e. $W$ remains constant. To keep the object's roundness factor constant, we also keep the interocular $b$ proportional to the object distance $H$ (\(b = \alpha H\)). The effect on perceived depth is shown in Fig.~\ref{fig:zoomingin}. We see that the depth magnification or roundness factor close to the screen plane remains equal to 1, as expected, but the depth of out-of-screen objects is distorted, and a closer analysis would show that the wide-interocular-zoomed-in configuration creates a \emph{puppet-theater effect}\index{Puppet-theater effect}, since farther objects have a larger image size than expected. This is one configuration where the roundness factor of the in-screen objects can be kept constant while changing the stereoscopic camera parameters, and in the next section we will show how to compute all the changes in camera parameters that give a similar result.

\subsection{Shape Distortions and the Depth Consistency Rule}
\label{sec:3-d-distortions}

If we want the camera setup and the display setup to preserve the shape of all observed objects up to a global \threeD{} scale factor, a first constraint is that there must be a linear relation between depths, so we must be in the canonical setup described by eq. (\ref{eq:6}). A second constraint is that the ratio between depth magnification and image magnification, called the \emph{roundness factor}, must be equal to 1. This means that the \emph{only} configuration with faithful depth reproduction is when the shooting and viewing geometries are \emph{homothetic} (i.e. there is a scale factor between the two geometries):
\begin{equation}
  \label{eq:9}
  \frac{W'}{W}=\frac{H'}{H} =\frac{b'}{b}.\qquad\text{(Homothetic configuration)}
\end{equation}

In general, the depth and space distortion is nonlinear: it can easily be shown that the perceived space is a homographic transform of the real space.
Shooting from farther away while zooming in with a bigger interocular doesn't distort (much) depth, as we showed in the previous section, and that's probably the right way to zoom in - the baseline should be proportional to the convergence distance. But one has to take care about the fact that the infinity plane in scene space may cause divergence.

More generally, we can introduce the \emph{depth consistency rule}:\index{Stereoscopic cinema!depth consistency rule}\index{Depth consistency rule|see{Stereoscopic cinema, depth consistency rule}} Objects which are close to the convergence plane in the real scene or close to the screen in projection space (\(Z = H\) or \(Z' = H'\)) should have a depth which is consistent with their apparent size, i.e. their roundness factor should be equal to 1.

The depth ratio between scene space and projection space close to the convergence plane can be computed from (\ref{eq:5}) as:
\begin{equation}
  \label{eq:10}
  \frac{\partial Z'}{\partial Z}(Z=H) = \frac{b}{HW}\frac{H'W'}{b'},
\end{equation}
and the apparent size ratio is simply \(\frac{W'}{W}\). Setting the ratio between both, i.e. the roundness factor, equal to 1 leads to the \emph{depth consistency rule}:
\begin{equation}
  \label{eq:11}
  \frac{b}{H} = \frac{b'}{H'}.\qquad\text{(Depth consistency rule)}
\end{equation}

One important fact arises from the depth consistency rule: for objects that are close to the convergence plane or the screen plane, screen size ($W'$) does not matter, whereas most spectators expect to get exaggerated \threeD{} effects when looking at a movie on a bigger screen. Be careful, though, that for objects farther than the screen, especially at $Z\rightarrow +\infty$, divergence may occur on bigger screens. Since neither $b$, $H$, or $b'$ can be changed when projecting a \threeD{} movie, the key parameter for the depth consistency rule is in fact screen distance, which dramatically influences the perceived depth: the bigger the screen distance, the higher the roundness factor will be. This means that if a stereoscopic movie, which was made to be seen in a movie theater from a distance of 16m, is viewed on a 3DTV from a distance of 4m, the roundness factor will be divided by 4 and the spectator will experience the classical \emph{cardboard effect}\index{Cardboard effect} where objects look flat.

What can we do to enforce the depth consistency rule, i.e. to produce the same \threeD{} experience in different environments, given the fact that $b’$ is fixed and $H'$ is usually constrained by the viewing conditions (movie theater vs. home cinema vs. TV)? The only possible solution consists in artificially changing the shooting parameters in post production using techniques that will be described in Sec.~\ref{cha:chang-shoot-geom}.

\subsection{Shooting With the Right Depth of Field}
\label{sec:depth-focus}

When disparities outside of the Percival zone of comfort are present, \citet{Ukai:2007} showed that vergence-accomodation conflicts that arise can be attenuated by reducing the depth of field: the ideal focus distance should be on the plane of convergence (i.e. the screen), and the depth of field should match the expected depth of field of the viewing conditions.  They even showed that objects that are out of this focus range are surprisingly better perceived in \threeD{} if they are blurred than if they are in focus.

We saw (Sec.~\ref{sec:depth-cues}) that the human eye depth of field in normal conditions is between 0.2D and 0.3D (diopters) -- let us say 0.2D to be conservative. This depth of field should be converted to a depth range in the targeted viewing conditions: \(Z'_{\min} = 1/\left(\frac{1}{H'}-0.2\right)\), \(Z'_{\max} = 1/\left(\frac{1}{H}+0.2\right)\) (if $Z'_{\max}$ is negative, then it is considered to be at infinity). Then, the perceived depth range should be converted to real depth range \(\left[Z_{min},Z_{max}\right]\), using the inverse of formula~(\ref{eq:5}), and the camera aperture should be computed from this depth range.

Unfortunately, the viewing conditions are usually not known at shooting time, or the movie has to be made for a various range of screen distances and sizes. We will describe a possible solution to this general case in Sec.~\ref{sec:changing-depth-focus}.

\subsection{Remaining Issues}
\label{sec:caveats}

Even when the targeted viewing conditions are known precisely, the shooting conditions are sometimes constrained: for example, when filming wild animals, a minimum distance may be necessary, resulting in using a wide baseline and a long focal length (i.e. a large \(\frac{b}{W}\)). The resulting stereoscopic movie, although it will have a correct roundness factor around the screen plane, will probably have strong divergence at infinity, since \(\frac{b'}{W'} < \frac{b}{W}\), which may look strange, even if the right depth of field is used.

Another kind of problem may happen, even with objects which are close to the screen plane: Psychophysics experiments showed that specular reflections may also be used as a shape cue, and more specifically as a curvature cue~\cite{Blake:1990,Blake:1991}.\index{Depth cues!specular reflections} Except when using a purely homothetic configuration, these depths cues will be inconsistent with other cues, even near the screen plane, and this effect will be particularly visible when using long focal lengths. This problem can probably not be overcome by changing the shooting geometry, and the best solution is probably to edit manually the specular reflections in post-production to make them look more natural, or to use the right makeup on the actors...

Inconsistent specular reflections can also be due to the use of a mirror rig (a stereoscopic camera rig using a half-silvered mirror to separate images for the left and right cameras): specular reflections are usually polarized by nature, and the transmission and reflection coefficients of the mirror depend on the polarization of incoming light. As a result, specular reflections have a different aspect in the left and right images.

\section{Lessons for Live-Action Stereoscopic Cinema from Animated \threeD{}}
\label{s:lessons}

Shooting live-action stereoscopic cinema has progressed over the last decade with the arrival of more versatile camera rigs.  Older stereo rigs had two fixed cameras, and the stereo extrinsic parameters -- baseline and vergence -- were changed by manual adjustment between shots.  Newer rigs allow dynamic change of the stereo parameters during a shot.  This still falls short of a final stage of versatility, which is the modification of the left- and right-eye images during post-production, to effectively change the stereo extrinsic parameters -- in other words, this ultimate goal is to use the shot footage as a basis for synthesizing left- and right-eye images with any required stereo baseline and vergence.

New technologies are bringing us closer to the goal of synthesizing left- and right-eye views with different stereo parameters during post-production.  But what are the benefits, and how will this create a better viewer experience?  To answer this question, in this section we look at current practices in animated stereoscopic cinema.  In animation, the creative team has complete control over camera position, camera motion, camera intrinsics, stereo extrinsics, and the \threeD{} structure of the scene.  This allows scope to experiment with the stereoscopic experience, including doing \threeD{} manipulations and distortions that do not correspond to any physical reality but which enhance viewer experience.  We describe four core techniques of animated stereoscopic cinema.  The challenge for live-action \threeD{} is to create new technologies so that these techniques can be applied to live-action footage as easily as they are currently applied to animation.
 
\subsection{Proscenium Arch or Floating Window} 
\label{s:floating-windows}
The proscenium arch or floating window was introduced earlier.  The simplest way to project stereoscopic imagery is to capture images from a left- and right- camera and then put those images directly onto the cinema screen.  This imposes a specific epipolar geometry on the viewer, and our eyes adjust to it.  Now consider the four-sided boundary of the physical cinema screen.  It also imposes epipolar constraints on the eyes (in fact these are epipolar constraints which are consistent with the whole rest of the physical world).  But note that there is no reason why the epipolar geometry imposed by the stereoscopic images will be consistent with the epipolar constraints associated with the screen boundary.  The result is conflicting visual cues.

The solution is to black-mask the two stereoscopic images.  Consider the basic \threeD{} animation setup, and two cameras observing a \threeD{} scene.  Now imagine between the camera and the scene a rectangular window, in \threeD{}, so that the cameras view the scene through the window, but the window is surrounded by a black wall where nothing is visible. This is the proscenium arch or floating window, a virtual \threeD{} entity interjected between the cameras and the scene.   The visual effect of the window is achieved by black-masking the boundaries of the left- and right- eye images.  Since this floating window is a \threeD{} entity that is consistent with the cameras and the rest of the \threeD{} scene, it does not give rise to conflicting visual cues.

\subsection{Floating Windows and Audience Experience}
\label{s:floating-experience}
Section~\ref{s:floating-windows} described a basic motivation for the floating window, motivated by comfort in the viewing experience.  There is a further use for the technique.  
Note that the window can be placed anywhere in the \threeD{} scene,  It can lie between the camera and the scene, part-way through the scene, or behind the scene.  This is not perceived explicitly but placing the \threeD{} scene behind the floating window relative to the audience produces a more subdued passive feeling, according to accepted opinion in the creative community.  Placing the scene in front of the floating window relative to the audience produces a more engaged active feeling.
Also note that the window does not need to be fronto-parallel to the viewer.  Instead it can be tilted in \threeD{} space.  Again the viewer is typically unaware of this consciously, but orientation of the floating window can produce subliminal effects e.g. a forward tilt of the upper part of the window can produce a looming feeling.

\subsection{Window Violations}
\label{s:window-violations}
The topic of floating windows leads naturally to another technique.  Consider again the \threeD{} setup - cameras, floating window, and \threeD{} scene.  First consider objects that are on the far side of the window from the cameras.  When they are center-stage, they are visible through the window.  As they move off-stage and are obscured by the surrounding wall of the virtual window.  This is all consistent from the viewpoint of the two stereoscopic cameras that are viewing the scene.  Now consider an object that lies between the cameras and the window.  While it's center-stage and visible in the two eyes, everything is fine.  As it moves off-stage, it intersects the view-frustrums created by each camera and the floating window, and nothing is visible outside the frustrum.  But this is inconsistent to the eye -- it's as though we are looking at someone in front of a doorway and their silhouette disappears as it passes over the view frustrums in the left- and right-eyes of the doorway which is to the rear of them.  

The solution to this problem is simply not to allow such violations.  This is straightforward in animation but in live-action stereoscopic cinema, of course, it requires the non-trivial recovery of the stereo camera positions and the \threeD{} scene to detect when this is happening, and avoid it.

\subsection{Multi-Rigging}
\label{s:multi-rigging}
So far, we considered manipulations of the left- and right- eye images that are consistent with a physical \threeD{} reality.  Multi-rigging, however, is a technique for creating stereoscopic images which produce a desired viewing experience, but could not have arisen from a physically correct \threeD{} situation.  Consider a scene with objects A and B.  The left camera is kept fixed but there are two right cameras, one for shooting object A and one for shooting object B.  Two different right eye images are generated, and they are then composited so that objects A and B appear in a single composite right eye image.  

Why do this?  Consider the case where A is close to the camera and B is far away.  In a normal setup, B will appear flat due to distance.    But by using a large stereo baseline for shooting object B, it is possible to capture more information around the occluding contour of the object, and give it a greater feeling of roundedness, even though it is placed more distantly in the scene.  Again this is straightforward for animated stereoscopic cinema, but requires \threeD{} capture of the scene, and the ability to modify stereo baseline at post-production time, to apply it to live-action stereoscopic cinema.

\section{Post-production of Stereoscopic Movies}
\label{cha:chang-shoot-geom}

When filming with stereoscopic cameras, even if the left and right views are rectified in post-production (Sec.~\ref{cha:elim-vert-disp}), there are many reasons for which the movie may not be adapted to given viewing conditions, among which:
\begin{itemize}
\item The screen distance and screen size are different from the ones the movie was filmed for, resulting in a different \emph{roundness factor} for objects close to the screen.
\item Because of a screen size larger than expected, the points at infinity cause eye divergence.
\item There were constraints on camera placement (as those that happen when filming sports or wildlife), which cause large disparities, or even divergence, on far-away objects.
\item The stereoscopic camera was not adjusted properly when filming.
\end{itemize}

As we will see during this section, changing the shooting geometry in post-production is theoretically possible, although the advanced techniques require high-quality computer vision and computer graphics algorithms to perform a process called view interpolation.  Due to the fact that \threeD{} information can be extracted from the stereoscopic movies, there are also a few other stereo-specific post-production processes that may be improved by using computer vision techniques.

\subsection{Eliminating Vertical Disparity: Rectification of Stereoscopic Movies}
\label{cha:elim-vert-disp}

Vertical disparity is one of the sources of visual fatigue in stereoscopic cinema, and it may come from misaligned cameras or optics, or from toed-in camera configurations, where vertical disparity cannot be avoided.

In Computer Vision, \threeD{} reconstruction from stereoscopic images is usually preceded by a transformation of the original images~\citep[Chap. 12]{Forsyth:2003}. This transformation, called \emph{rectification}\index{Rectification} is a \twoD{} warp of the images that aligns matching points on the same $y$ coordinate in the two warped images. The rectification of a stereoscopic pair of images is usually preceded by the computation of the \emph{epipolar geometry}\index{Epipolar geometry} of the stereoscopic camera system. Knowing the epipolar geometry, one can map a point in one of the images, to a line or curve in the other image which is the projection of the optical ray issued from that point onto that image. The rectification process transforms the epipolar lines or curves into horizontal lines, so that stereoscopic matching by computer vision algorithms is made easier.

It appears that this rectification process is exactly what we need to eliminate vertical disparity in stereoscopic movies, so that all the available results from Computer Vision on computing the epipolar geometry \cite{Hartley:2000,Fitzgibbon:2001,Micusik:2003,Barreto:2005,Steele:2006} or on the rectification of stereoscopic pairs \cite{Hartley:1999,Loop:1999,Fusiello:2000,Abraham:2005,Wu:2007,Zhou:2006,Cheng:2009} will be useful to accurately eliminate vertical disparity.

However, the rectification of a single image pair acquired in a laboratory with infinite depth of field for stereoscopic matching, and rectification of a stereoscopic movie that will be presented to spectators is not exactly the same task. The requirements of a rectification method designed for stereoscopic cinema are:
\begin{enumerate}
\item it should be able to work without knowing anything from the stereoscopic camera parameters, because this data is not always available, and may be lost in the film production pipeline, whereas images are usually not lost;
\item it should require no calibration pattern or grid, since the wide range of camera configurations and optics would require too many different calibration grid sizes: all the computation should be made from the images themselves (we call it \emph{blind} rectification);
\item the aspect ratio of rectified images should be as close as possible to the aspect ratio of the original images;
\item the rectified image should fill completely the frame: no ``unknown'' or ``black'' area can be tolerated;
\item the rectification of the whole movie should be smooth (jitter or fast-varying rectification transforms will create shaky movies) so that the rectification parameters should either be computed for a whole shot, or it should be slowly varying over time;
\item the camera parameters (focal length, focus...) and the rig parameters (interocular, vergence..) may be fixed during a shot, but they could also be slowly varying;
\item the images may have artistic qualities that are difficult to handle for computer vision algorithms: lack of texture, blur, saturation (white or black), sudden change in illumination, noise, etc.
\end{enumerate}

Needless to say, very few rectification methods fulfill these requirements, and the best solution will probably have to take the best out of the cited methods in order to give acceptable results. \citet{Cheng:2009} recently proposed a method to rectify stereo sequences with varying camera motions and zooming effects, while keeping the image aspect ratio, but there is still room for improvements, and there will probably be many other publications on this subject in the near future.

A proper solution would probably contain the following ingredients:
\begin{itemize}
\item A method for epipolar geometry computation and rectification that takes into account nonlinear distortion \cite{Barreto:2005,Steele:2006,Abraham:2005}: even if cinema optics are close-to-perfect, images have a very high resolution, and small nonlinear distortions with an amplitude of a few pixels may happen at short focal lengths.
\item A multi-scale feature detection and matching method, which will be able to handle blurred or low-texture areas in the images.
\item Proper parameterization of the rectification functions, so that only rectifications that conserve the aspect ratio are allowed. Methods for panorama stitching reached this goal by using the camera rotation around its optical center as a parameter -- this may be a good direction.
\item Temporal filtering, which is not an easy task since the filtered rectifications must also satisfy the above constraints.
\end{itemize}

\subsection{Shifting and Scaling the Images}
\label{sec:shifting-images}

Image shifting is a process already used by \citet{Spottiswoode:1952}, in particular to achieve \emph{continuity in space} during shot transitions. The human brain is actually not very sensitive to stereoscopic shifting: if two fixed rectified images are shown in stereo to a subject, and the images are shifted slowly, the subject will not notice the change in shift, and surprinsingly the scene will not appear to move closer or farther as would be expected. However, shifting the images modifies the eye vergence and not the accomodation, and thus may break the vergence-accomodation constraints and cause visual fatigue: when images are shifted, one must verify that the disparities are still within Percival's zone of comfort (which depends on viewing geometry).

Image shifting is mostly used for ``softening the cuts'': If the disparity of the main area of interest changes abruptly from one shot to the other, the human eyes may take some time to adjust the vergence, and it will cause visual fatigue or even alter the oculo-motor system, as shown by~\citet{Emoto:2005}. As explained by \citet{Spottiswoode:1952}, shifting the images is one solution to this problem.

One way to make a smooth transition between shots is the following: let us say the main subject of shot 1 is at disparity $d_1$, and the main subject of shot 2 is at disparity $d_2$ (they do not have to be at the same position in the image). About one second before the transition, the images from shot 1 are slowly shifted from \(d_0=0\) to \(d_0=\frac{d_2-d_1}{2}\). When the transition (cut or fade) happens, the disparity of the main subject in shot 1 is \(d=\frac{d1+d2}{2}\). During the transition, the first image of shot 2 is presented with a shift of \(d_0=\frac{d1-d2}{2}\), so that the disparity of the main subject is also \(d=\frac{d1+d2}{2}\). After the transition, the disparity continues to be slowly shifted, in order to arrive at a null shift (\(d_0=0\)) about one second after the transition.

During this process, care must be taken not only to stay inside Percival's zone of comfort, but also to avoid divergence in the areas in focus (the divergence threshold depends on screen size).

Another simple post-production process consists in scaling the images. Scaling is equivalent to changing the width $W$ of the convergence plane, and allows reframing the scene by panning simultaneously both rescaled images.

There is no quality loss when shifting or scaling the images, except the one due to resampling the original images. In particular, no spatial or temporal artifact may appear in shifted or scaled sequences, whereas they may happen in the processes described hereafter.

\subsection{View Interpolation, View Synthesis, and Disparity Remapping}
\label{sec:view-interpolation}

\subsubsection{Definitions and Existing Work}
\label{sec:descr-techn}

Image shifting and scaling cannot solve all the issues caused by having different (i.e. non-homothetic) shooting and viewing geometries. Ideally, we would like to be able to change two parameters of the shooting geometry in order to get homothetic setups: the camera interocular $b$ and the distance to the convergence plane $H$. This section describes the computer vision and computer graphics techniques which can be used to achieve these effects in stereoscopic movie post-production. The results that these methods have obtained so far are not on par with the picture quality requirements of the stereoscopic cinema, but this research field is progressing constantly, and these requirements will probably be met within the next few years.

\begin{figure}[htbp]
  \centering
  \begin{tabular}{|c|c|c|}
    \hline
    Method & Shooting geometry & Viewing geometry \\
    \hline
    \hline
    \begin{sideways}\parbox{6cm}{\centering View interpolation}\end{sideways} &
    \includegraphics[scale=0.5]{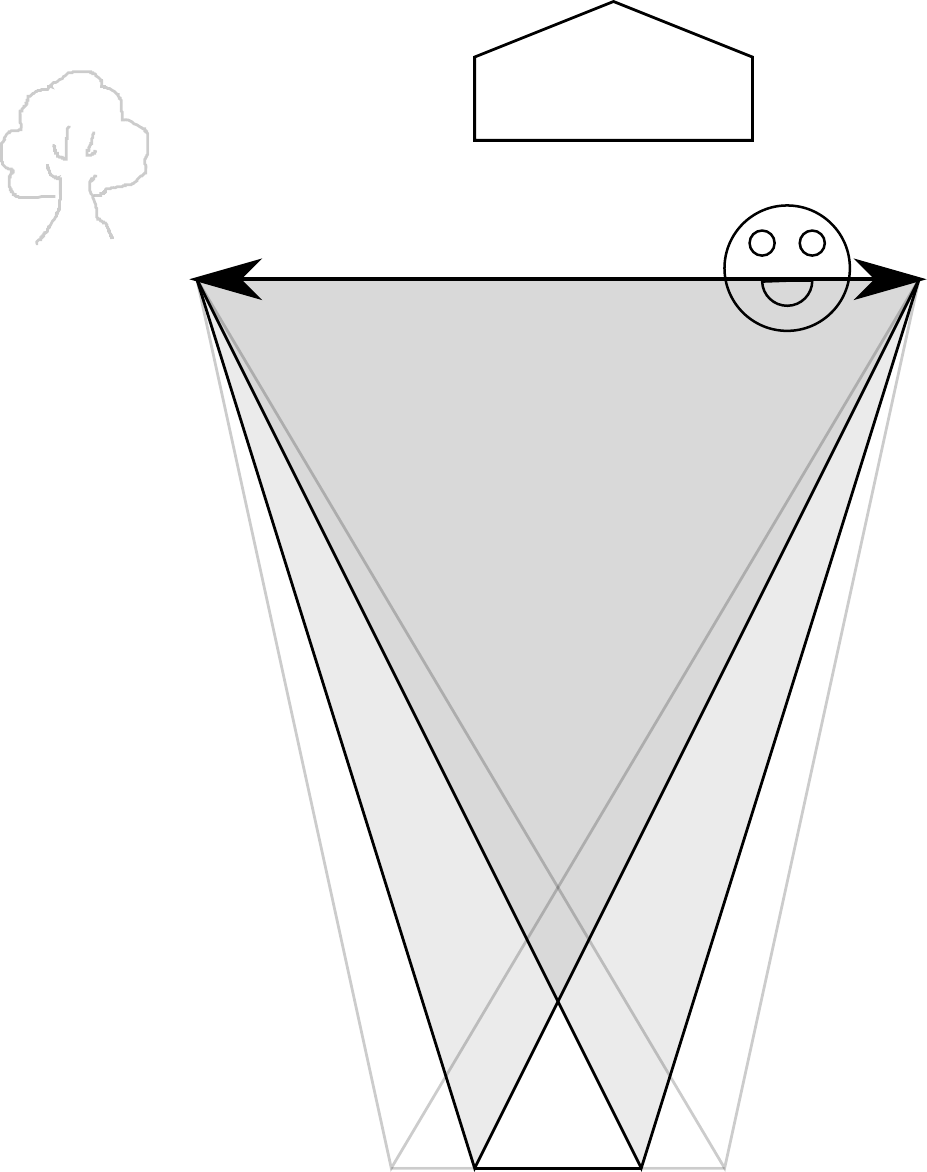} &
    \includegraphics[scale=0.5]{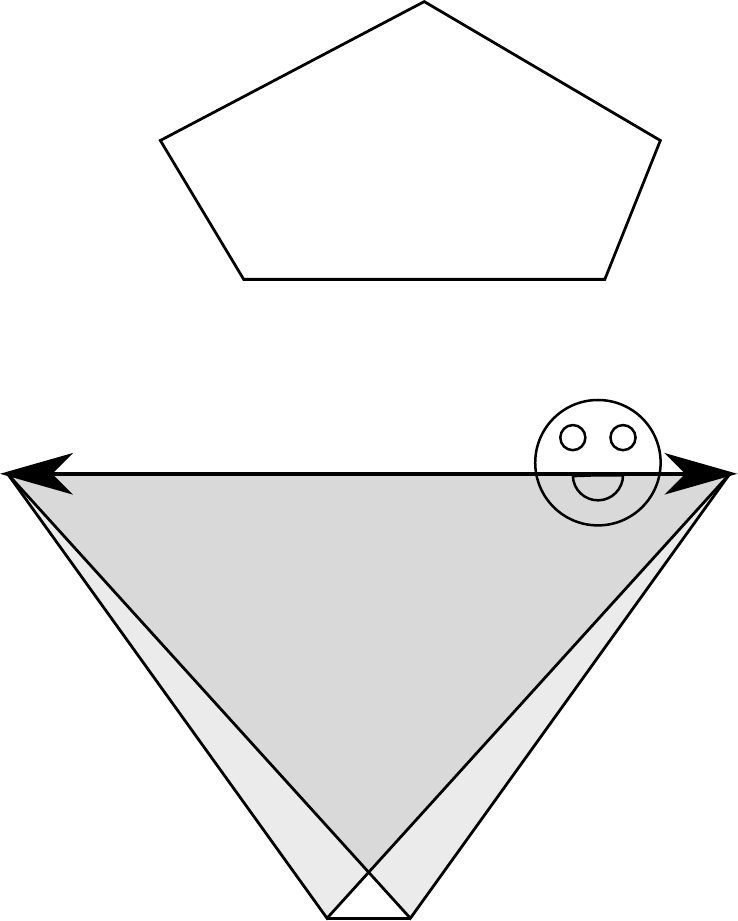} \\
    \hline
    \begin{sideways}\parbox{6cm}{\centering View synthesis}\end{sideways} &
    \includegraphics[scale=0.5]{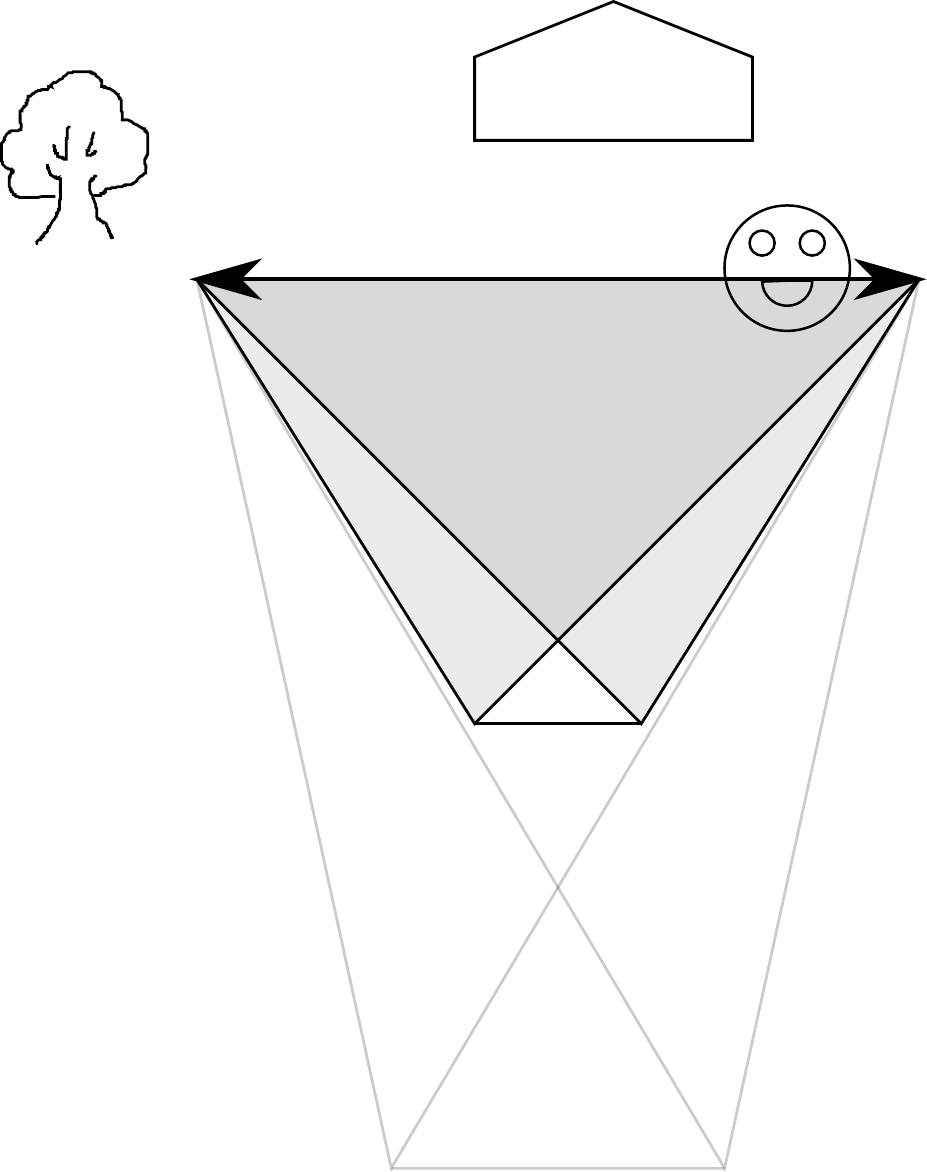} &
    \includegraphics[scale=0.5]{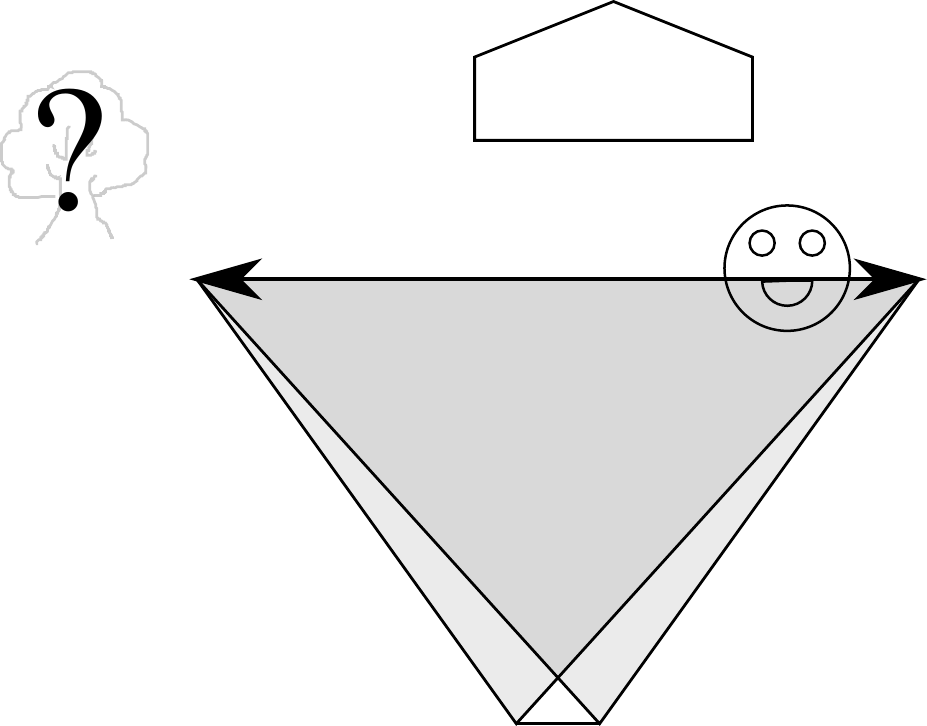} \\
    \hline
    \begin{sideways}\parbox{6cm}{\centering Disparity remapping}\end{sideways} &
    \includegraphics[scale=0.5]{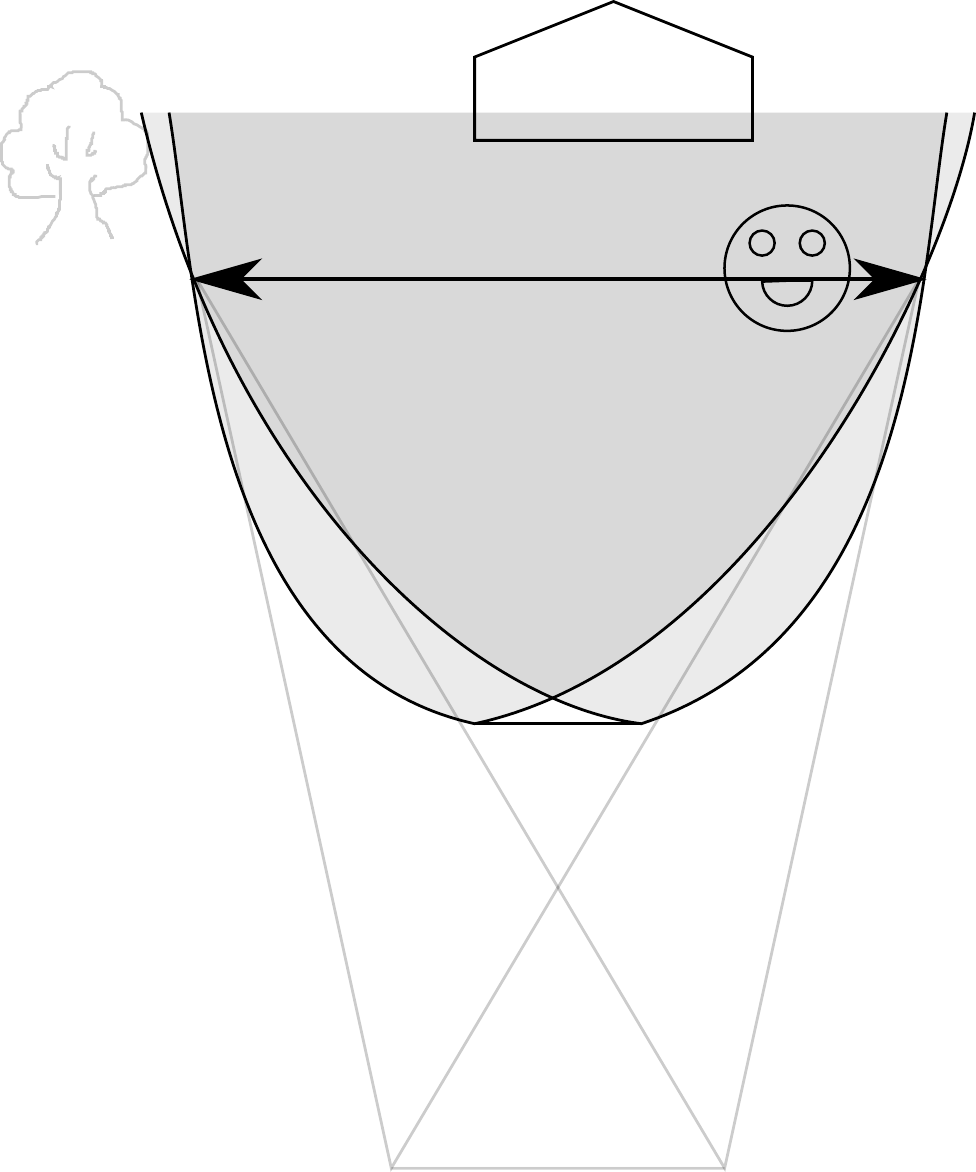} &
    \includegraphics[scale=0.5]{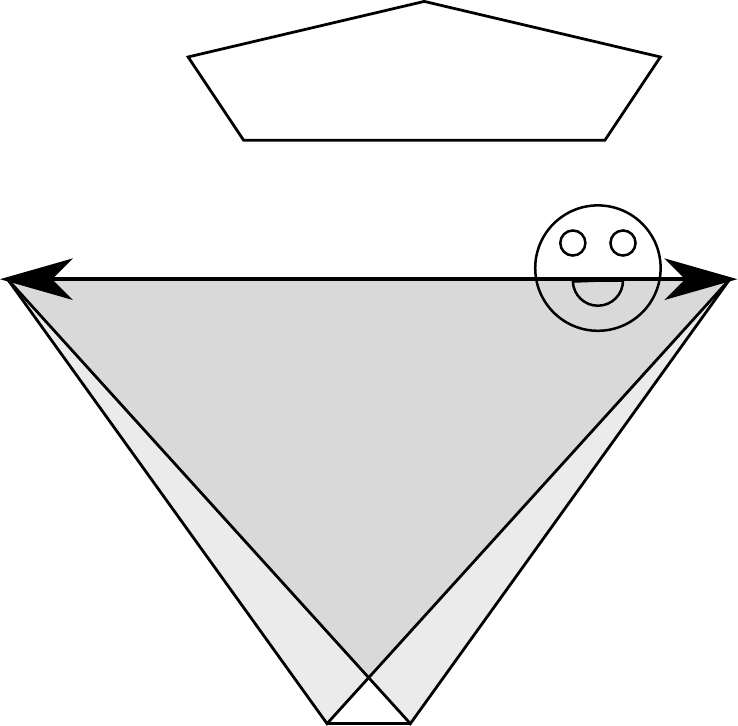} \\
    \hline
  \end{tabular} 
  \caption{Changing the shooting geometry in post-production (original shooting geometry is drawn in light gray in the left column). All methods can preserve the roundness of objects in the screen plane, at some cost for out-of-screen objects: view interpolation distorts dramatically their depth and size; view synthesis generates a geometrically correct scene, but it is missing a lot of scene information about objects that cannot be seen in the original images; disparity remapping may preserve the depth information of the whole scene, but the apparent width of objects is not preserved, resulting in puppet-theater effect.}
  \label{fig:geometry-interp}
\end{figure}

The first thing we would like to do is change the camera interocular $b$, in order to follow the depth consistency rule \(\frac{b}{H} = \frac{b'}{H'}\).  To achieve this, we use \emph{view interpolation}\index{View interpolation}, a technique that takes as input a set of synchronized images taken from different but close viewpoints, and generates a new view \emph{as if} it were taken by a camera placed at a specified position between the original camera positions. It is different from a well-known post-production effect called \emph{retiming}, which generates intermediate images between two consecutive images in time taken by the same camera, although some people have been successfully using retiming techniques to do view interpolation if the cameras are close enough. As shown in the first line of Fig.~\ref{fig:geometry-interp}, although the roundness factor of objects near the screen plane is well preserved, depth and size distortions are present and are what would be expected from the interpolated camera setup: far objects are heavily distorted both in size and depth, and divergence may happen at infinity.

If we also want to change the distance to screen, we have to use \emph{view synthesis}\index{View synthesis}. It is a similar technique, where the cameras may be farther apart (they can even surround the scene), and the synthesized viewpoint can be placed more freely. View synthesis usually requires a shooting setup with at least a dozen of cameras, and may even use hundreds of cameras. It is sometimes used in free-viewpoint video, when the scene is represented as a set of images, not as a textured \threeD{} mesh. The problem is that we usually only have two cameras, and for view synthesis to work, at least all parts of the scene that would be visible in the synthesized viewpoint must be also visible in the original viewpoints. Unfortunately, we easily notice (second line of Fig.~\ref{fig:geometry-interp}) that many parts of the scene that were not visible in the original viewpoints become visible in the interpolated viewpoints (for example the tree in the figure). Since invisible parts of the scene cannot be invented, view synthesis is clearly not the right technique to solve this problem.

What we propose is a mixed technique between view interpolation and view synthesis, that preserves the global visibility of objects in the original viewpoints, but does not produce depth distortion or divergence: \emph{disparity remapping}. In disparity remapping (third line of Fig.~\ref{fig:geometry-interp}), we apply a nonlinear transfer function to the disparity function, so that perceived depth is proportional to real depth (the scale factor is computed at the convergence plane). In practice, it consists in shifting all the pixels in the image that have a given disparity by the disparity value that would be perceived in the viewing geometry for an object placed at the same depth. That way, divergence may not happen, since objects that were at a given distance from the convergence plane will be projected at the same distance, up to a fixed scale factor, and thus points at infinity are effectively projected at infinity. However, since the object image size is not changed by disparity remapping, there will still be some kind of puppet-theater effect, since far-away objects may appear bigger in the image than they should be. Of course, any disparity remapping function, even one that does not conserve depth, could be used to get special effects, to obtain effects similar to the multi-rigging techniques used in animation (Sec.~ \ref{s:multi-rigging}).

There has been already a lot of work on view interpolation~\cite{Zitnick:2004,Criminisi:2007,Park:2004,Taguchi:2008,Bleyer:2009,Xiong:2009,Zitnick:2006,Hasinoff:2006,Kilner:2006,Woodford:2007,Xiong:2007,Wang:2008,Rogmans:2009}, and disparity remapping as we defined it uses the exact same tools. They have solved the problem with various levels of success, and the results are constantly improving, but none could definitely get rid of the two kinds of artifacts that plague view interpolation:
\begin{itemize}
\item spatial artifacts, which are usually ``phantom'' objects or surfaces that appear due to specular reflections, occluded areas, blurry or non-textured areas, semi-transparent scene components (although many recent methods handle semi-transparency in some way).
\item temporal artifacts, which are mostly ``blinking'' effects that happen if the spatial artifact are not temporally consistent and seem to ``pop up'' at each frame.
\end{itemize}

\subsubsection{Asymmetric Processing}
\label{sec:asymm-proc}

The geometry changes leading to view interpolation shown in Fig.~\ref{fig:geometry-interp} are symmetric, i.e. the same changes are made to the left and right view. Interpolating both views may alter the quality and create artifacts and both views, resulting in a lower-quality stereoscopic movie. However, this may not be the best way to change the shooting geometry by interpolation.

\citet{Stelmach:2000} showed that the quality of the binocular percept is dominated by the sharpest image. Consequently, the artifacts could be reduced by interpolating only one of the two views, and perhaps smoothing the interpolated view if it contains artifacts. However, a more recent study by \citet{Seuntiens:2006a} contradicts the work of \citet{Stelmach:2000}, and shows that the perceived-quality of an assymetrically-degraded image pair is about the average of both perceived qualities when one of the two views is degraded by a very strong JPEG compression. Still, they admit that ``asymmetric coding is a valuable way to save bandwidth, but one view must be of high quality (preferably the original) and the compression level of the coded view must be within acceptable range'', so that for any kind of assymetric processing of a stereo pair (such as view interpolation), one has to determine what is this ``acceptable range'' to get a perceived quality which is better than with symmetric processing. \citet{Kooi:2004} also note that interocular blur suppression, i.e. fusing an image with a blurred image, requires a few more seconds than with the original image pair.  \citet{Gorley:2008} devised another metric for stereoscopic pair quality, which is based on comparing areas around SIFT features. This proved to be more consistent with perceived quality measured by human subjects than the classical PSNR (Peak Signal-to-Noise Ratio) measurement used in \twoD{} image quality measurements.  Further experiments should be led on the perception of assymetrically-coded stereoscopic sequences to get a better idea of the allowed amount of degradation in one of the images.

One could argue that the best-quality image in an asymmetrically-processed stereoscopic sequence should be the one corresponding to the dominant eye\footnote{About seventy per cent of people prefer the right eye, and thirty per cent prefer the left, and this is correlated to right- or left-handedness, but many people are cross-lateral (e.g. right-handed with a left dominant eye).}, but this means that interpolation should depend on the viewer, which is not possible in a movie theater or in consumer 3DTV. \citet{Seuntiens:2006a} also studied the effect of eye dominance on the perception of assymetrically-compressed images, and their conclusion is that eye dominance has no effect on the quality perception, which means that it may be possible to do assymetric processing while remaining user-independent. 


\subsection{Changing the Depth of Field}
\label{sec:changing-depth-focus}

We have seen in sections \ref{sec:verg-accom-confl} and \ref{sec:depth-focus} that not only the shooting geometry should be adapted to the display, but also the depth of field: the in-focus plane should be at the screen depth (i.e. at zero disparity), and the depth of field should follow the human depth of field (i.e. 0.2D to 0.3D), so that there is no visual fatigue due to horizontal disparity limits or vergence-accomodation conflicts.

Having a limited depth of field may seem limited at first, but \citet{Ukai:2007} showed that \threeD{} is actually perceived even for blurred objects, and this also results in less visual fatigue when these objects are out of the theoretical focus range.  Another big issue with stereoscopic displays is the presence of crosstalk\index{Stereoscopic cinema!crosstalk}. Lenny Lipton claims \cite{Lipton:2001}: ``Working with a digital projector, which contributes no ghosting artifact, proves that the presence of even a faint ghost image may be as important as the accomodation/convergence breakdown.''.  Actually, blurring the objects that have high disparities should also strongly reduce that ghosting effect, so that changing the depth of field may have the double advantage of diminishing visual fatigue and reducing crosstalk.

In animated \threeD{} movies, the depth of field can be artificially changed, since images with infinite depth of field can be generated, together with a depth map: \cite{Kakimoto:2007,Bertalmio:2004,Lin:2007}. In live-action stereoscopic cinema, the depth of field can be reduced using similar techniques, given a depth map extracted from the live stereoscopic sequence. Increasing the depth of field is theoretically possible, but would enhance dramatically the image noise.

\subsection{Dealing With the Proscenium Rule}
\label{sec:break-prosc-rule}

The proscenium rule states that the stereoscopic sequence should be seen at though it were happening behind a virtual arch formed by the stereoscopic borders of the screen, called the proscenium arch.  When an object appears in front of the proscenium, and it touches either the left or the right edges of the screen, then the proscenium rule is violated, because part of the object becomes not visible in both eyes, whereas it should be (see Sec.~\ref{s:floating-windows} and Fig.~\ref{fig:proscenium}). When the object touches the top or bottom borders, the stereoscopic scene is still consistent with the proscenium arch, and some even argue that in this case the proscenium appears bended towards the object~\citep[Chap. 5]{Mendiburu:2009}.

Since the proscenium arch is a virtual window formed by the \threeD{} reconstruction of the left and right stereoscopic image borders, it can be virtually moved forward or backward.  Moving the proscenium forward is the most common solution, especially when using the screen borders as the proscenium would break the proscenium rule, both in live-action stereoscopic movies and in animated \threeD{} (Sec.~\ref{s:window-violations}). It can also be a narrative element by itself, as may be learned from animated \threeD{} (Sec.~\ref{s:floating-experience}). The left and right proscenium edges may even have different depths, without the audience even noticing it (this effect is called ``floating the stereoscopic window'' by \citet{Mendiburu:2009}).

The proscenium arch is usually moved towards the spectator by adding a black border on the left side of the right view and on the right side of the left view. However, on display devices with high rates of crosstalk\index{Stereoscopic cinema!crosstalk}, moving the proscenium may generate strong artifacts on the screen left and right borders, where one of the views is blackened out.  One way to avoid the crosstalk issues is to use a semi-opaque mask instead of a solid black mask, as suggested and used by stereographer Phil Streather.  Other alternative solutions solutions may be tried, such as blurring the edges, or a combination of blurring and intensity attenuation.

\subsection{Compositing Stereoscopic Scenes and Relighting}
\label{sec:comp-ster-scen}

Compositing\index{Compositing} consists in combining several image sources, either from live action or computer-generated, to obtain one shot that is ready to be edited. It can be used to insert virtual objects in real scenes, real objects in virtual scenes, or real objects in real scenes. The different image sources should be acquired using the same camera parameters. For live-action footage, this does usually mean that either match-moving or motion-controlled cameras should be used to get the optical camera parameters (internal and external). Matting masks then have to be extracted for each image source: masks are usually automatically-generated for computer-generated images, and several solutions exist for live-action footage: automatic (blue screen / green screen / chroma key), or semi-automatic (rotoscoping / hand drawing). The image sources are then combined together with matting information to form the final sequence.  In non-stereoscopic cinema, if there is no camera motion, the compositing operation is often \twoD{} only, and a simple transform (usually translation and scaling) is applied to each \twoD{} image source.

The \threeD{} nature of the stereoscopic cinema imposes a huge constraint on compositing: not only the image sources must be placed correctly in each view, but the \threeD{} geometry must be consistent across all image sources. In particular, this means that the shooting geometry (Sec.~\ref{cha:keep-prop-pick}) must be the same, or must be adjusted in post-production to match a reference shooting geometry (Sec.~\ref{sec:view-interpolation}). If there is a \threeD{} camera motion, it can be recovered using the same match-moving techniques as in \twoD{} cinema, which may take into account the rigid link between left and right cameras to recover a more robust camera motion.

The matting stage could also be simplified in stereoscopic cinema: the disparity map (which gives the horizontal disparity value at each pixel in the left and right rectified views) or the depth map (which can be easily computed from the disparity map) can be used as a Z-buffer to handle occlusions automatically between the composited scenes. The disparity map representation should be able to handle transparency, since the pixels at depth discontinuities should have two depths, two colors, and a foreground transparency value~\citep{Zitnick:2004}. Recent algorithms for stereo-based \threeD{} reconstruction automatically produce both a disparity map and matting information, which can be used in the compositing and view interpolation stages~\citep{Xiong:2007,Taguchi:2008,Bleyer:2009,Xiong:2009,Sizintsev:2010}. When working on a sequence that has to be modified using view interpolation or disparity remapping, it should be noted that the interpolated disparity map can be obtained as a simple transform of the disparity map computed from the original rectified sequence, and this solution should be preferred to avoid the \threeD{} artifacts caused by running a stereo algorithm on interpolated images (more complicated transforms happen in the case of view synthesis).


Getting consistent lighting \index{Relighting} between the composited image sources can be difficult: even if the two scenes to be composed were captured or rendered using the same lighting conditions, the mutual lighting (or radiosity) will be missing from the composited scene, as well as the cast shadows from one scene to the other.

If lighting consistency cannot be achieved during the capture of real scenes (for example if one of the shots comes from archive footage), the problem becomes much more complicated: it is very difficult to change the lighting of a real scene, even when its approximate \threeD{} geometry can be recovered (by stereoscopic reconstruction in our case), because that \threeD{} reconstruction will miss the material properties of the \threeD{} surfaces, i.e. how each surface reacts to incident light, and how light is back-scattered from the surface, and the \threeD{} texture and normals of the surface are very difficult to estimate from stereo. These material properties are contained in the BRDF (Bi-Directional Reflectance Function), which has a complicated general form with many parameters, but can be simplified from some classes of materials (e.g. isotropic, non-specular...).

The brute force solution to this problem would be to acquire the real scene under all possible lighting conditions, and then select the right one from this huge data set. Hopefully, only a subset of all possible lighting conditions is necessary to be able to recompose any lighting condition: A simple set of lighting conditions (the light basis) is used to capture the scene, and more complicated lighting conditions can be computed by linear combination of the images taken under the light basis. This is the solution adopted by Debevec et al. for their \emph{Light Stage} setup \citep{Chabert:2006,Debevec:2002}.

\subsection{Adding Titles or Subtitles}
\label{sec:adding-titles-or}

Whereas adding titles or subtitles in \twoD{} is rather straightforward (the titles should not overlap with important information in the image, such as faces or text), it becomes more problematic in stereoscopic movies.

Probably the most important constraint is that titles should never appear inside or behind objects in the scene: this would result in a conflict between depth (the object is in front of the title) and occlusion (the text occludes the object) which will cause visual fatigue in the long run.

Another constraint is that titles should appear at the same depth as the focus of attention, because re-adjusting the vergence may take as much as one second, and cause visual fatigue~\citep{Emoto:2005}.  The position of the title within the image is less important, since the oculomotor system has a well-trained ``scanning'' function for looking at different places at the same depth. If the display device has too much crosstalk, then a better idea may be to find a place within the screen plane which is not occluded by objects in the scene (because they are highly contrasted, out-of-screen titles may cause double images in this case).

A gross disparity map produced by automatic \threeD{} reconstruction should be enough to help satisfy these two constraints, the only additional information that has to be provided is \emph{where} is the person who is speaking. Then, the position and depth can be computed from the disparity map and the speaker location: The position can be at the top or the bottom of the image, or following the speaker's face like a speech ballon, and the depth should either be the screen depth (at zero disparity) or the depth of the speaker's face.

\section{Conclusion}

This chapter has reviewed the current state of understanding in stereoscopic cinema.  It has discussed perceptual factors, choices of camera geometry at time of shooting, and a variety of post-production tools to manipulate the the \threeD{} experience.  We included lessons from animated stereoscopy to illustrate how the creative process proceeds when there is complete control over camera geometry and \threeD{} content, indicating useful goals for stereoscopic live-action work.  Looking to the future, we anticipate that \threeD{} screens will become pervasive in cinemas, in the home as  \threeD{} TV, and in hand-held \threeD{} displays.  A new market in consumer stereoscopic photography has also made an appearance in 2009, as Fuji released the W1 Real \threeD{} binocular-stereoscopic camera plus auto-stereoscopic photo frame.  All of this activity promises exciting developments in stereoscopic viewing and in the underlying technology of building sophisticated \threeD{} models of the world geared towards stereoscopic content.

\scriptsize

\bibliographystyle{spbasic}
\bibliography{geoofstereocinebib,paulfredbib.bib}


\end{document}